\documentclass[letterpaper,journal]{IEEEtran}
\usepackage{amsmath,amsfonts}
\usepackage{algorithmic}
\usepackage{algorithm}
\usepackage{array}
\usepackage{textcomp}
\usepackage{stfloats}
\usepackage{url}
\usepackage{verbatim}
\usepackage{graphicx}
\usepackage{capt-of}
\usepackage{cite}
\usepackage{nccrules}
\usepackage{xspace}
\usepackage{etoolbox}
\usepackage{booktabs}
\usepackage{multirow}
\usepackage[table]{xcolor}
\usepackage{hyperref}
\hypersetup{hidelinks}
\usepackage{makecell}

\makeatletter
\let\IEEEorigmakecaption\@makecaption
\long\def\@makecaption#1#2{%
\ifx\@captype\@IEEEtablestring
  \footnotesize\bgroup\par\@IEEEtabletopskipstrut
  \setbox\@tempboxa\hbox{\normalfont\footnotesize
    {#1.}\nobreakspace\nobreakspace #2}%
  \ifdim\wd\@tempboxa>\hsize
    \setbox\@tempboxa\hbox{\normalfont\footnotesize
      {#1.}\nobreakspace\nobreakspace}%
    \parbox[t]{\hsize}{\normalfont\footnotesize\noindent
      \unhbox\@tempboxa#2}%
  \else
    \hbox to\hsize{\normalfont\footnotesize\box\@tempboxa\hfil}%
  \fi
  \par\addvspace{0.5\baselineskip}\egroup
  \@IEEEtablecaptionsepspace
\else
  \IEEEorigmakecaption{#1}{#2}%
\fi}
\makeatother
\renewcommand{\tablename}{Table}

\definecolor{myorange}{RGB}{245,156,74}
\definecolor{mygray}{gray}{0.4}        
\definecolor{myred}{RGB}{197,41,114}
\definecolor{mygreen}{RGB}{76,120,80}

\hypersetup{  
	colorlinks = true,     %
	linkcolor = blue,      %
	urlcolor = blue,       %
	citecolor = blue,      %
	linkbordercolor = red
}

\definecolor{curveblue}{RGB}{31, 119, 180}
\definecolor{clblue}{RGB}{222, 234, 246}
\newcommand{\cc}[1]{\cellcolor{clblue!50}{#1}}

\usepackage{pifont}
\newcommand{\xmark}{\ding{55}}
\newcommand{\cmark}{\ding{51}}

\RequirePackage{xspace}
\makeatletter
\DeclareRobustCommand\onedot{\futurelet\@let@token\@onedot}
\def\@onedot{\ifx\@let@token.\else.\null\fi\xspace}

\def\eg{\emph{e.g}\onedot} 
\def\Eg{\emph{E.g}\onedot}
\def\ie{\emph{i.e}\onedot}

\makeatother

\def\method{UniRestorer$++$}

\usepackage{amsmath,amsfonts,bm}

\def\eqref#1{equation~\ref{#1}}

\def\1{\bm{1}}

\DeclareMathAlphabet{\mathsfit}{\encodingdefault}{\sfdefault}{m}{sl}
\SetMathAlphabet{\mathsfit}{bold}{\encodingdefault}{\sfdefault}{bx}{n}

\begin{document}

\title{UniRestorer++: Scalable Universal Image\\Restoration with Multi-Granularity\\Degradation Modeling}

\author{Jingbo Lin, Zhilu Zhang, Fan Li, Renjing Pei, Wenbo Li, \\
Hang Xu, Hongzhi Zhang, and Wangmeng Zuo}

\maketitle

\begin{abstract}
	Universal image restoration aims to handle diverse degradations within a single model. Despite rapid progress, a reliable scaling strategy remains underexplored. 
    The conventional strategy of simply increasing backbone depth and width can exhibit performance saturation, with optimization conflicts across restoration objectives remaining a potential bottleneck. 
    Degradation-aware methods allocate specialized capacity, but finer specialization requires more accurate degradation estimation and becomes vulnerable to estimation errors, especially on out-of-distribution and real-world inputs.
	To this end, we propose UniRestorer$++$, a scalable universal image restoration framework driven by multi-granularity degradation modeling. Specifically, we construct a degradation hierarchy for model scaling by progressively refining the degradation group and introducing group-specific restoration experts, thereby coupling capacity growth with degradation-specific specialization. To ensure reliable scaling, degradation-estimation-based routing selects a root-to-leaf path within the hierarchy, while granularity-estimation-based routing selects an appropriate level along that path, balancing fine-grained specialization with robustness against degradation-estimation errors. Furthermore, spatially dense degradation representation maps enable region-aware expert routing to adaptively handle spatially non-uniform corruptions.
    We develop the CDD-50 benchmark, encompassing 50 degradation categories ranging from single-degradation to penta-degradation. 
    Extensive experiments show approximately log-linear PSNR trends over the evaluated model-size range, with stronger scaling gains on out-of-distribution and real-world data than vanilla MoE scaling. UniRestorer$++$ achieves state-of-the-art performance across comprehensive benchmarks. It also demonstrates zero-shot generalization and few-shot transfer capability on real-world images.
	The code and pre-trained models will be publicly available at \url{https://github.com/mrluin/UniRestorer_plusplus}.
\end{abstract}

\section{Introduction}
\label{sec:introduction}

\IEEEPARstart{I}{mage} restoration aims to recover high-quality visual content from degraded observations, supporting applications in computational photography~\cite{Feijoo_2025_CVPR}, medical imaging~\cite{Yan_Region_MICCAI2024}, remote sensing~\cite{11093597}, autonomous driving~\cite{11093933}, and security surveillance~\cite{liao2025gmmoe}. 
Early methods were typically tailored to individual degradations, \eg, DnCNN~\cite{DnCNN} for denoising and SRCNN~\cite{SRCNN} for super-resolution. As degradations diversified and neural architectures advanced, research expanded from task-specific methods to task-agnostic restoration backbones, including MPRNet~\cite{MPRNet}, SwinIR~\cite{SwinIR}, Restormer~\cite{Restormer}, NAFNet~\cite{NAFNet}, and X-Restormer~\cite{xrestormer}. 
More recently, \emph{universal image restoration} has emerged to address multiple restoration tasks and diverse degradation types within a single unified model.

\begin{figure}[!t]
	\centering
	\includegraphics[width=\columnwidth]{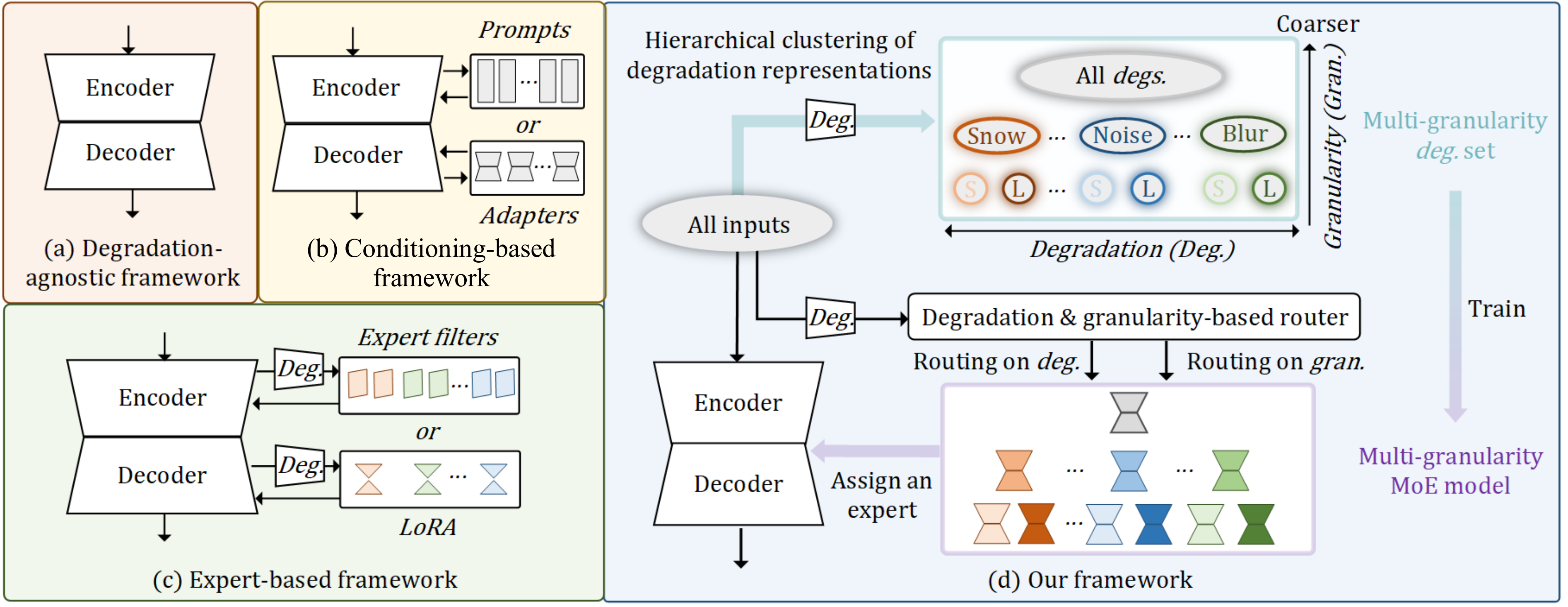}
\caption{Representative universal image restoration frameworks.
 (a) Degradation-agnostic methods use a shared backbone without explicit degradation conditioning.
 Degradation-aware methods guide restoration through conditioning (b) or expert-based processing (c) that routes features toward degradation-specific experts.
 These representative methods typically use a fixed degradation granularity, which can make them sensitive to degradation-estimation errors.
 (d) \method~proposes multi-granularity degradation modeling, expanding model capacity by introducing finer degradation-specific restoration experts. Degradation-estimation-based routing determines a hierarchical path, while granularity-estimation-based routing selects an expert level along that path, balancing degradation-specific specialization with robustness to degradation-estimation errors.
 }
\label{fig:introduction_type_of_framework}
\end{figure}

Existing universal restoration methods can be broadly categorized into degradation-agnostic and degradation-aware methods.
Degradation-agnostic methods jointly train a shared backbone on data across all tasks (see Fig.~\ref{fig:introduction_type_of_framework}(a)), but conflicting restoration objectives (such as denoising and deblurring) can hinder restoration quality. 
Degradation-aware methods use degradation information to guide restoration and can be further divided into conditioning-based and expert-based approaches.
Conditioning-based methods use learnable prompts~\cite{PromptIR, Transweather, OneRestore, mioir}, degradation representations~\cite{AirNet, IDR}, or adapters~\cite{MPerceiver} to modulate a shared restoration backbone (see Fig.~\ref{fig:introduction_type_of_framework}(b)). 
Expert-based methods allocate degradation-specific capacity through specialist models~\cite{RestoreAgent, grids}, selective filters~\cite{ADMS, WGWSNet}, low-rank adaptation (LoRA) experts~\cite{lorair}, routing features toward specialized pathways (see Fig.~\ref{fig:introduction_type_of_framework}(c)).

\begin{figure*}[!t]
    \centering
    \begin{minipage}[t]{0.67\textwidth}
        \vspace{0pt}
        \centering
        \includegraphics[width=\linewidth]{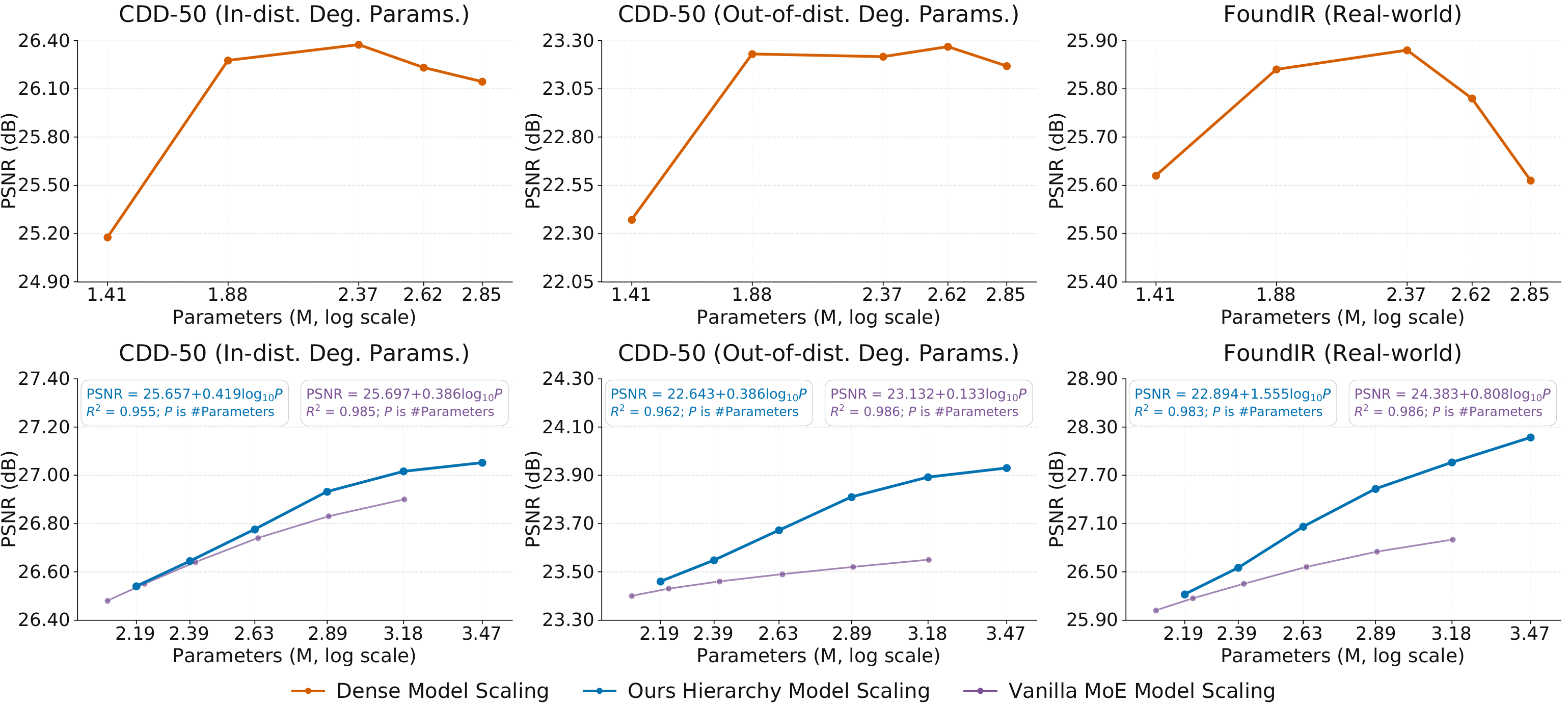}
        \captionof{figure}{
        	Model scaling evaluation on our CDD-50 and FoundIR~\cite{foundir} datasets. The equations give empirical $\operatorname{PSNR}=a+b\log_{10}P_{\mathrm{total}}$ fits for \textcolor[HTML]{0072B2}{our hierarchy model scaling} and \textcolor[HTML]{7A5195}{vanilla MoE model scaling}. See Section~\ref{sec:scalability} for more details.}
        \label{fig:scaling_up_params_psnr_ours}
    \end{minipage}
    \hfill
    \begin{minipage}[t]{0.315\textwidth}
        \vspace{0pt}
        \centering
        \includegraphics[width=\linewidth]{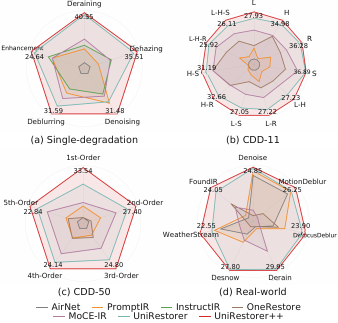}
        \captionof{figure}{Universal restoration performance on single-degradation tasks, CDD-11, CDD-50, and real-world datasets.}
        \label{fig:left}
    \end{minipage}
\end{figure*}

Despite rapid progress in universal restoration, existing methods still exhibit a noticeable performance gap compared to specialist models, particularly when confronted with an expanding and diverse degradation space. 
Drawing inspiration from advances in large language models~\cite{LLM,T5} and large-scale vision models~\cite{scaling_vit,SwinV2,DINOv3}, scaling up the capacity of universal restoration models represents a promising direction to close this gap and potentially surpass specialist models.
However, simply increasing the depth and width of a dense restoration network yields limited gains in our experiments (see the upper part of Fig.~\ref{fig:scaling_up_params_psnr_ours}), as it does not explicitly allocate capacity to different degradation subspaces, leading to conflicts among different restoration objectives.
To reduce parameter interference, degradation-aware methods use conditioning or expert specialization to adapt restoration processing to different degradation subspaces. 
Nevertheless, the efficacy of these methods critically hinges on degradation estimation, where the finer parameter specialization required for model scaling demands higher estimation accuracy.
For out-of-distribution and real-world degradations, degradation-estimation errors can activate mismatched parameters and compromise the benefit of increased specialization (see `Vanilla MoE Scaling' in the lower part of Fig.~\ref{fig:scaling_up_params_psnr_ours}).
\textit{\textbf{In this paper, we argue that a scalable universal restoration framework must ensure that capacity expansion effectively translates into both predictable performance and generalization gains. Achieving this goal fundamentally requires scaling up the model while maintaining robustness against degradation-estimation errors.}
}

To address these challenges, we present \method, a scalable universal image restoration framework driven by multi-granularity degradation modeling, as illustrated in Fig.~\ref{fig:introduction_type_of_framework}(d).
\textbf{First}, we construct a degradation hierarchy to govern the capacity scaling of universal restoration models. Each subsequent hierarchy level refines the degradation space of the preceding level into finer groups and introduces group-specific restoration experts for the resulting subspaces. The expansion of model capacity is thereby achieved by increasing the depth of this hierarchy.
\textbf{Second}, while deeper hierarchies provide more specialized capacity, they can also amplify the consequence of degradation-estimation errors. To enable reliable scaling, we employ a multi-granularity expert routing mechanism that balances expert specialization with robustness against degradation-estimation errors.
Specifically, fine-grained experts specialize in narrow degradation subspaces, whereas coarse-grained experts cover broader distributions to improve robustness under ambiguous estimates. 
Degradation-estimation-based routing determines a root-to-leaf path within the hierarchy, while granularity-estimation-based routing selects an appropriate level along that path.
For instance, an ambiguous estimate of heavy snow can be routed to a coarse-grained general-snow-removal expert, whereas a confident estimate can activate a fine-grained heavy-snow-removal expert. 
\textbf{Third}, existing methods generally use an image-level degradation representation to condition restoration, without explicitly selecting different experts for individual spatial locations. However, haze density, blur patterns, and many real-world degradations often vary spatially. To address this issue, our method predicts spatially dense degradation representation maps for region-aware expert routing, allowing different experts to be selected for distinct regions in an image.

We conduct extensive experiments across multiple evaluation settings.
We develop the CDD-50 benchmark, which encompasses 50 distinct degradation categories ranging from single corruptions to complex mixtures of up to five simultaneous degradations.
Experimental results demonstrate that \method~consistently establishes state-of-the-art performance across in-distribution, out-of-distribution, and real-world degradation benchmarks. 
Crucially, the proposed hierarchical model scaling strategy exhibits approximately log-linear PSNR trends over the evaluated model-size range, with steeper fitted slopes than vanilla mixture-of-experts (MoE) scaling on out-of-distribution CDD-50 and FoundIR, as shown in Fig.~\ref{fig:scaling_up_params_psnr_ours}.
By scaling the parameter capacity up to 2.9B, we establish the largest known fidelity-oriented universal restoration model, which demonstrates fine zero-shot generalization and few-shot transfer capabilities on real-world images.

A preliminary version of this work was published in the Proceedings of ICLR 2026~\cite{lin2026unirestorer}. 
Methodologically, the present manuscript extends the original multi-granularity routing mechanism into a scalable universal restoration framework by establishing a hierarchical model scaling method and generalizing global routing to spatially adaptive expert routing. 
Empirically, this work constructs a comprehensive CDD-50 evaluation benchmark, while providing investigations of scaling behavior, out-of-distribution generalization, and few-shot transferability.

The contributions are summarized as follows:

\begin{itemize}
	\item We propose a hierarchy model scaling method for universal image restoration, which progressively refines the degradation partition and assigns restoration experts to the resulting subspaces.
	\item We propose a multi-granularity expert routing mechanism, which balances expert specialization with robustness against degradation-estimation errors.
	\item We present a region-aware expert routing scheme, allowing local regions to dynamically activate distinct restoration experts for spatially non-uniform degradations.
	\item We establish the CDD-50 benchmark encompassing 50 degradation categories. Extensive experiments characterize the empirical log-linear scaling trend over the evaluated model-size range and demonstrate superior performance in universal restoration, real-world zero-shot generalization, and few-shot transferability.
\end{itemize}

\section{Related Work}
\label{sec:related_work}
\subsection{Image Restoration for Specific Degradations}
Early image restoration methods developed architectures tailored to individual degradations, \eg, DnCNN~\cite{DnCNN} for denoising, SRCNN~\cite{SRCNN} for super-resolution, DeblurGAN-v2~\cite{DeblurGANv2} for deblurring, PReNet~\cite{PReNet} for deraining, Dehazeformer~\cite{Dehazeformer} for dehazing, and DesnowNet~\cite{Snow100K} for desnowing. Advances in deep architectures produced general-purpose restoration backbones, including MPRNet~\cite{MPRNet}, SwinIR~\cite{SwinIR}, NAFNet~\cite{NAFNet}, Restormer~\cite{Restormer}, and X-Restormer~\cite{xrestormer}. 
These backbones can be trained separately for different restoration tasks and provide specialist baselines.

\subsection{Universal Image Restoration}
\noindent\textbf{Overview.}~ 
Universal image restoration aims to process multiple known or unknown degradation types within a single model. Existing approaches fall into two categories: degradation-agnostic and degradation-aware.
Degradation-agnostic methods share all model weights and learn cross-task knowledge implicitly from mixed training data, which suffer from restoration-task conflicts.
Degradation-aware methods instead estimate degradation representations (DRs) and use them as priors, following the paradigm of blind image restoration~\cite{dzsr_pami, dzsr}. Within this category, methods can be further divided into conditioning-based and expert-based approaches according to how the degradation priors guide restoration.

\noindent\textbf{Conditioning-based Methods.}~ 
Conditioning-based degradation-aware methods~\cite{PromptIR, IDR, Transweather, AirNet, mioir, MPerceiver} integrate prompting~\cite{prompt}, adapters~\cite{adapter}, or feature modulation~\cite{SRMD} into a shared restoration backbone. Such mechanisms can support parameter-efficient fine-tuning (PEFT), but their use alone does not imply that the backbone is frozen. TransWeather~\cite{Transweather} and AirNet~\cite{AirNet}, \eg, learn degradation-related embeddings that modulate intermediate features for different corrupted inputs. PromptIR~\cite{PromptIR} combines prompt learning with a Restormer backbone~\cite{Restormer}, whereas IDR~\cite{IDR} integrates task-oriented prior hubs through learnable principal component analysis and dynamic routing. MiOIR~\cite{mioir} combines explicit and adaptive prompts, while OneRestore~\cite{OneRestore} learns text-aligned prompts for composite-degradation restoration. 
However, the degradation-specific processing they induce can be sensitive to errors in the conditioning representation, especially over a wide and complex degradation space.

\noindent\textbf{Expert-based Methods.}~
Mixture-of-experts models have demonstrated conditional processing capability in multi-task learning~\cite{mod_squad, mthl, uniperceivermoe}. Expert-based degradation-aware methods implement specialization at different scales, from lightweight degradation-specific modules to full restoration networks. \Eg, WGWSNet~\cite{WGWSNet} and ADMS~\cite{ADMS} learn specific filters on top of a shared backbone. DaAIR~\cite{DaAIR} uses a degradation-aware learner in a low-rank regime, Instruct-IPT~\cite{InstructIPT} learns low-rank task-specific weight offsets, and LoRA-IR~\cite{lorair} uses low-rank experts. MoCE-IR~\cite{moce} learns experts with different computational complexities and receptive fields. At the full-model level, GRIDS~\cite{grids} groups degradations and trains restoration networks for the resulting groups. The related RestoreAgent system~\cite{RestoreAgent} instead uses a multimodal language model to select restoration tools and their execution order.

Despite this progress, degradation-aware approaches typically select prompts or experts at a fixed degradation granularity.
Although such selection enables degradation-specific processing, it relies on degradation-estimation accuracy and is therefore vulnerable to degradation-estimation errors.
Moreover, many such approaches use image-level degradation cues without explicitly selecting different experts for individual spatial locations. Spatially varying feature responses alone do not provide this form of region-aware expert selection.

\subsection{Scaling Universal Image Restoration Models}
Larger models and training sets have improved capability and generalization in natural language processing and high-level vision tasks. Beyond dense scaling of model depth and width, sparse MoE models enlarge model capacity while activating only a subset of parameters for each input~\cite{sparse_gated_moe, gshard, switch_transformer}. 

Comprehensive model scaling remains underexplored in universal image restoration. Existing work has improved restoration capacity through higher-capacity backbones~\cite{Restormer, xrestormer, chen2023hat}, pretraining on broader degradation spaces~\cite{DegAE, kong2024gir}, or larger high-quality datasets~\cite{HQ_50K}. GenLV~\cite{genlv} studies scalable unified modeling across a wide spectrum of low-level vision tasks using three dense model variants from $39$M to $211$M parameters.
Scaling laws have characterized empirical relationships between model capacity and performance in language models~\cite{s_base} and high-level vision models~\cite{scaling_vit,revisiting_scaling_laws}. Routed-model scaling laws additionally distinguish total capacity from active computation~\cite{s_base}. These studies do not directly address how parameters should be organized across heterogeneous image degradations. Consequently, whether universal image restoration exhibits a predictable scaling relationship and how parameter organization affects this relationship remain underexplored.

\section{Proposed Method}
\label{sec:proposed_method}

\begin{figure*}[!t]
	\centering
	\includegraphics[width=\textwidth]{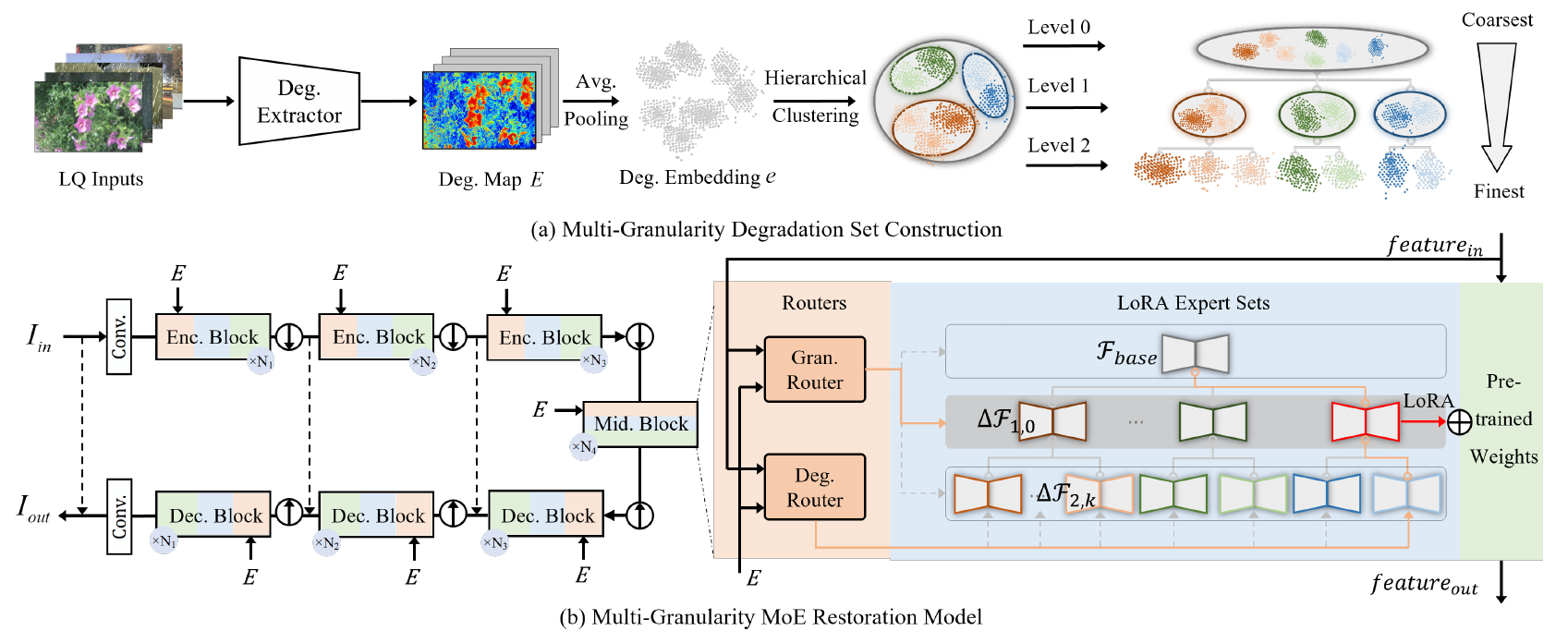}
    \vspace{-6mm} 
	\caption{
        Overview of \method. Globally pooled degradation representations (DRs) are hierarchically clustered to construct a multi-granularity degradation set, while dense DR maps are retained for region-aware expert routing. At each restoration block, the degradation-estimation-based and granularity-estimation-based routers are conditioned on both dense DR priors and the current restoration feature, enabling expert routing to adapt throughout the network. Their joint decision produces an expert assignment at each spatial location.
    }
    \label{fig:framework}
\end{figure*}

We first introduce the basic formulation of model scaling, then compare dense model scaling and vanilla MoE scaling to motivate our hierarchy model scaling strategy.
We then detail the components supporting our hierarchy model scaling strategy.
The overview of our method is shown in Fig.~\ref{fig:framework}. 
We first learn dense degradation representations (DRs) and hierarchically cluster them to construct a multi-granularity degradation set.
Then, we accordingly train a multi-granularity mixture-of-experts (MoE) restoration model.
To keep increasingly fine-grained specialization robust to uncertain degradation estimates, the hierarchy retains experts at multiple granularities, allowing ambiguous inputs to be handled by broader degradation groups. The mechanism for selecting an appropriate granularity is introduced later.
As a result, the dense DR maps and routing strategy enable adaptively assigning inputs to appropriate experts across spatial locations and throughout the network.

\subsection{Preliminaries}
\label{sec:preliminaries}

\noindent\textbf{Basic Formulation of Model Scaling.}~
Let $\mathcal{E}$ denote the expected mean squared restoration error under a given evaluation distribution. 
Under ideal optimization without multi-task interference, the scaling behavior of a monolithic neural network with parameter count $P$ is characterized by the floor-plus-power-law relation~\cite{revisiting_scaling_laws}:
\begin{equation}
	\mathcal{E}(P)
	\simeq \mathcal{E}_{\mathrm{floor}} + c P^{-\alpha},
	\label{equ:dense_scaling_model}
\end{equation}
where $c > 0$ denotes the reducible error coefficient, $\alpha > 0$ represents the fundamental power-law scaling exponent, and $\mathcal{E}_{\mathrm{floor}} \geq 0$ denotes the asymptotic error floor determined by intrinsic task ambiguity and evaluation constraints. 
The power-law exponent $\alpha$ serves as the foundational metric of parameter efficiency, quantifying the rate of error decay per order of magnitude of parameter expansion.

\noindent\textbf{Mixture-of-Experts.}~
A mixture-of-experts (MoE) architecture consists of an expert set $\{\mathcal{F}_a\}_{a=1}^{N_{\mathrm{exp}}}$ and a routing network $\mathcal{G}$. The expert set contains $N_{\mathrm{exp}}$ parallel modules, \eg, convolutional or feed-forward layers. Given an input $\bm{x}$ and its routing representation $\bm{v}$, $\mathcal{G}_a(\bm{v})$ denotes the contribution of the $a$th expert $\mathcal{F}_a(\bm{x})$ to the MoE output $\bm{y}$,
\begin{equation}
	\bm{y}=\sum_{a=1}^{N_{\mathrm{exp}}}
	\mathcal{G}_a(\bm{v})\mathcal{F}_a(\bm{x}).
	\label{eq:moe}
\end{equation}
One sparse routing formulation is a noisy top-$1$ router parameterized by $\bm{W}_{\mathrm{gate}}$ and $\bm{W}_{\mathrm{noise}}$. It applies softmax to the noisy routing logits and retains only the expert with the largest routing probability,
\begin{equation}
	\begin{aligned}
		\mathcal{G}(\bm{v})
		&=
		\operatorname{TopK}\!\Big(
		\operatorname{Softmax}\!\big(
		\bm{v}\bm{W}_{\mathrm{gate}}
		\\[-1mm]
		&\qquad+
		\bm{\epsilon}\odot
		\operatorname{Softplus}(\bm{v}\bm{W}_{\mathrm{noise}})
		\big);K\Big),
	\end{aligned}
	\label{eq:noisy_topk}
\end{equation}
where $\bm{\epsilon} \sim\mathcal{N}(\bm{0},\bm{I})$ is standard Gaussian noise, $\odot$ denotes element-wise multiplication, $\operatorname{TopK}(\cdot;1)$ sets all but the largest element to zero, and $\operatorname{Softplus}(\cdot)$ is a smooth approximation to ReLU.

\subsection{Scaling Universal Image Restoration Models}
\label{sec:scaling_rationale}
\label{sec:scaling_mechanism}

\noindent\textbf{Dense Model Scaling Strategy.}~
The conventional dense model scaling strategy expands the parameter count $P$ by uniformly increasing the depth and width of a shared restoration backbone. 
Because all degradation tasks share an unpartitioned parameter manifold, scaling $P$ intensifies gradient interference and optimization conflicts among contradictory restoration objectives (such as high-frequency detail synthesis versus low-pass noise smoothing). 
Consequently, the effective scaling exponent decays rapidly as model size increases. 
Beyond a critical parameter scale, multi-task optimization conflicts dominate the optimization trajectory, driving the effective exponent to zero and eventually into negative values, which manifests empirically as performance saturation and subsequent collapse.

\noindent\textbf{Vanilla MoE Model Scaling Strategy.}~
Sparse Mixture-of-Experts (MoE) architectures decouple parameter interference by routing visual representations to specialized sub-networks. 
Let $P_{\mathrm{total}}$ denote the total parameter capacity of the network, and let $P_{\mathrm{act}}$ denote the active parameter count. 
Under optimal expert assignment, the expected restoration error is governed by~\cite{s_base}:
\begin{equation}
	\mathcal{E}_{\mathrm{MoE}}
	\simeq \mathcal{E}_{\mathrm{floor}}
	+ c_1 P_{\mathrm{act}}^{-\alpha_1} P_{\mathrm{total}}^{-\alpha_2}
	+ c_2 P_{\mathrm{act}}^{-\alpha_3},
	\label{equ:moe_scaling_model}
\end{equation}
where $c_1, c_2, \alpha_1, \alpha_2, \alpha_3 > 0$ are structural scaling constants. 
Under a fixed active computation budget $P_{\mathrm{act}}$, the term $\alpha_2 > 0$ represents the theoretical upper-bound exponent achievable through total parameter expansion. 
In universal restoration, vanilla MoE scaling partitions the degradation space into increasingly fine, disjoint clusters governed by a flat router operating at a single granularity. 
However, as $P_{\mathrm{total}}$ expands, the candidate expert space proliferates, rendering degradation estimation and expert selection progressively more difficult. 
Under ambiguous or out-of-distribution (OOD) degradations, escalating routing misassignments introduce estimation noise that degrades the theoretical capacity exponent $\alpha_2$ into a substantially attenuated effective exponent $\alpha_{\mathrm{eff},\mathrm{vanilla}} \ll \alpha_2$, severely constraining the scaling return of total parameters.

\noindent\textbf{Hierarchy Model Scaling Strategy.}~
To mitigate the impact of degradation estimation error on the scaling exponent, we structure the degradation space into a sequence of nested partitions $\Pi_0 \prec \Pi_1 \prec \cdots \prec \Pi_{L-1}$, where $\prec$ denotes partition refinement and $L$ denotes the total number of hierarchy levels. 
Under a fixed active budget $P_{\mathrm{act}}$, we model the effective restoration error for a scaling strategy $\kappa \in \{\mathrm{vanilla}, \mathrm{hier}\}$ via an effective scaling exponent $\alpha_{\mathrm{eff},\kappa}$,
\begin{equation}
	\mathcal{E}_\kappa(P_{\mathrm{total}}) 
	\simeq \mathcal{E}_{\mathrm{floor}} + C_\kappa P_{\mathrm{total}}^{-\alpha_{\mathrm{eff},\kappa}},
	\label{equ:hierarchical_scaling_model}
\end{equation}
where $C_\kappa > 0$ denotes the strategy-dependent effective error coefficient, and $\alpha_{\mathrm{eff},\kappa} > 0$ characterizes the empirical scaling efficiency in the presence of routing dynamics. 
In vanilla MoE scaling, escalating routing errors under distribution shifts severely penalize the system, causing $\alpha_{\mathrm{eff},\mathrm{vanilla}}$ to diminish sharply. 
In contrast, our hierarchy model scaling strategy enables ambiguous or OOD inputs to fall back dynamically to coarser ancestral experts. 
By safeguarding expert routing against estimation uncertainties across all scales, hierarchy model scaling preserves the theoretical potential of capacity growth, sustaining a high and stable scaling exponent $\alpha_{\mathrm{eff},\mathrm{hier}} > \alpha_{\mathrm{eff},\mathrm{vanilla}}$ across both in-distribution and out-of-distribution conditions.

For images normalized to $[0,1]$, $-10\log_{10}\mathcal{E}_\kappa$ is the PSNR computed from the expected MSE. Under Eq.~(\ref{equ:hierarchical_scaling_model}), a nonzero error floor generally makes this quantity nonlinear in $\log_{10}P_{\mathrm{total}}$, its slope approaches $10\alpha_{\mathrm{eff},\kappa}$ only when the reducible error dominates the floor.
The reported average PSNR instead averages image-wise logarithmic errors and is generally different from the PSNR of the expected MSE. We therefore fit the reported average PSNR directly, using the following log-linear relation as an empirical summary over the evaluated model-size range:
\begin{equation}
	\overline{\operatorname{PSNR}}_\kappa(P_{\mathrm{total}})
	\simeq a_\kappa^{\mathrm{fit}} + b_\kappa^{\mathrm{fit}} \log_{10} P_{\mathrm{total}},
	\label{equ:predicted_log_scaling}
\end{equation}
where $\overline{\operatorname{PSNR}}_\kappa$ denotes the reported average PSNR, $a_\kappa^{\mathrm{fit}}$ is the fitted intercept, and $b_\kappa^{\mathrm{fit}}$ measures the fitted PSNR gain per tenfold increase in total parameters.
Under the negligible-floor approximation, interpreting $b_\kappa^{\mathrm{fit}}\simeq10\alpha_{\mathrm{eff},\kappa}$ additionally requires the gap between the two PSNR aggregation conventions to remain approximately constant with model size.
We use the measured PSNR and these fitted slopes to compare our hierarchy model scaling and vanilla MoE model scaling over the evaluated range
(see Fig.~\ref{fig:scaling_up_params_psnr_ours} and Section~\ref{sec:scalability}).

\subsection{Degradation Hierarchy Construction}
\noindent\textbf{Dense Degradation Representation Extractor.}~
Aiming to enable routing adaptive to spatially varying degradations, we first learn to extract dense DR maps.
Inspired by DA-CLIP~\cite{daclip}, we use text to describe both degradation type and degree and train the DR extractor by aligning DRs with their text embeddings.
Using an image-level CLIP~\cite{CLIP} representation, particularly after resizing or cropping high-resolution inputs, can discard local degradation cues.
We therefore use a ResNet-18-based~\cite{ResNet} dense DR extractor $\mathcal{D}$. Given a low-quality image $\mathbf{X}$, $\mathcal{D}$ predicts a dense DR map $\bm{E}$ with $p$ channels,
\begin{equation}
	\bm{E}=\mathcal{D}(\mathbf{X}),
	\label{equ:dense_dr}
\end{equation}
where $p$ denotes the DR dimension.
We construct fine-grained degradation captions by refining each degradation type into degree intervals,
\eg, for Gaussian noise, we divide $\sigma_{\mathrm{noise}}\in[0,\,50]$ into $[0,\,15)$, $[15,\,35)$, and $[35,\,50]$, described as `small noise', `medium noise', and `large noise', respectively.
We encode these captions into text embeddings that serve as the supervision targets for DR learning.
To provide pixel-wise supervision, we apply CutMix~\cite{yun2019cutmix} between degraded images with different degradation types and degrees, producing spatially varying inputs and their pixel-wise type-and-degree labels.
We optimize $\mathcal{D}$ using a pixel-wise contrastive loss that aligns each local DR with the text embedding of its corresponding degradation,
\begin{equation}
	\mathcal{L}_{\mathrm{pix}}
	=\frac{1}{|\Omega_d|}\sum_{r\in\Omega_d}
	\ell_{\mathrm{con}}\!\left(
	\bm{E}_{r},\bm{t}_{q(r)}
	\right),
	\label{equ:pixel_contrastive}
\end{equation}
where $\ell_{\mathrm{con}}$ denotes the contrastive loss, $q(r)$ is the type-and-degree label at spatial location $r$, $\bm{t}_{q(r)}$ is its text embedding, and $\Omega_d$ is the set of spatial locations in the dense DR map. The pixel-wise labels are aligned with this map for supervision. Consequently, $\bm{E}$ preserves local degradation characteristics instead of collapsing them into a single image-level representation.
After training, $\mathcal{D}$ is applied to all $N_{\mathrm{tr}}$ low-quality training samples, yielding dense DR maps $\{\bm{E}_n\}_{n=1}^{N_{\mathrm{tr}}}$.
We apply global average pooling for multi-granularity degradation set construction,
\begin{equation}
	\bm{e}_n=\operatorname{GAP}(\bm{E}_n)\in\mathbb{R}^{p},
	\label{equ:global_dr}
\end{equation}
while retaining dense maps $\bm{E}_n$ for routing.

\noindent\textbf{Hierarchical Degradation Clustering.}~
We recursively apply K-means clustering~\cite{deep_clustering} to organize the pooled DRs into a tree with $L$ levels. Let $\mathcal{S}_{i,j}$ denote the set of training-sample indices in the $j$-th DR group at level $i\in\{0,\ldots,L-1\}$, and let $J_i$ be the number of groups at that level. Thus, $\Pi_i=\{\mathcal{S}_{i,j}\}_{j=1}^{J_i}$. Level $0$ contains the single root group $\mathcal{S}_{0,1}=\{1,\ldots,N_{\mathrm{tr}}\}$, with $J_0=1$, and level $L-1$ is the finest.
For each parent group $\mathcal{S}_{i-1,m}$, we recursively partition its samples by K-means,
\begin{equation}
	\begin{aligned}
		&\{\bm{u}_{i,m,\nu}\}_{\nu=1}^{b_{i-1,m}}\\
		&\quad\leftarrow\operatorname{K\text{-}means}
		\big(\{\bm{e}_n\mid n\in\mathcal{S}_{i-1,m}\};\,b_{i-1,m}\big),
	\end{aligned}
\end{equation}
where $i\in\{1,\ldots,L-1\}$, $m$ indexes parent groups, $b_{i-1,m}$ denotes the number of resulting child groups, and $\bm{u}_{i,m,\nu}$ is the $\nu$-th child centroid of parent group $\mathcal{S}_{i-1,m}$. Each sample is assigned to its nearest child centroid. The resulting child groups are indexed by $j$ across level $i$, giving $J_i=\sum_{m=1}^{J_{i-1}}b_{i-1,m}$ groups at that level.
Recursively partitioning all parent groups yields non-overlapping DR groups at progressively finer granularities.
These nested DR groups form the degradation hierarchy introduced in Section~\ref{sec:scaling_mechanism}. Extending it by one level refines the degradation partition and defines the subspaces to which new restoration experts are assigned.

\subsection{Multi-Granularity MoE Restoration Model} %
\label{sec:multi_granularity_moe}
\noindent\textbf{Hierarchical Restoration Experts.}~ 
We implement the group-specific restoration parameters introduced in Section~\ref{sec:scaling_mechanism} as low-rank experts on a shared restoration backbone $\mathcal{F}_{\mathrm{base}}$. The shared backbone captures knowledge common to the degradation space. For each DR group $\mathcal{S}_{i,j}$ at level $i\geq1$, we learn a LoRA expert $\Delta\mathcal{F}_{i,j}$ on top of the frozen base weights, allocating additional capacity to that degradation subspace.
For block $s$, $\mathcal{F}_{\mathrm{base}}^{(s)}$ denotes the base block, and $\Delta\mathcal{F}_{i,j}^{(s)}$ denotes the LoRA residual of expert $(i,j)$.
We use rank-$8$ LoRA experts in our experiments.
Given prediction $\hat{\mathbf{Y}}$ and target $\mathbf{Y}$, the loss is,
\begin{equation}
	\ell_1=\|\hat{\mathbf{Y}}-\mathbf{Y}\|_1.
	\label{equ:restoration_loss}
\end{equation}
Coarse experts cover broad degradation subspaces and provide robust restoration under ambiguous estimations, whereas fine-grained experts specialize in narrower subspaces when degradation estimation is reliable. Adding a hierarchy level therefore expands specialized capacity without discarding the broader restoration capability represented by ancestor experts.

\noindent\textbf{Multi-Granularity Expert Routing.}~ 
Multi-granularity expert routing includes two complementary parts: degradation-estimation-based routing determines a root-to-leaf path, and granularity-estimation-based routing selects an appropriate level along that path.
We abbreviate the corresponding degradation-estimation-based router and granularity-estimation-based router as DER and GER, respectively.
Following uncertainty learning~\cite{Chang_2020_CVPR}, we model the DR at each spatial location as a Gaussian distribution with diagonal covariance to capture fine-grained degradation-estimation uncertainty. From $\bm{E}$, two estimation heads predict,
\begin{equation}
	\begin{aligned}
		\bm{E}_{d}&=\phi_{\mu}(\bm{E}),\\
		\bm{E}_{g}&=\phi_{\sigma^2}(\bm{E}),
	\end{aligned}
	\label{equ:degradation_distribution}
\end{equation}
where $\phi_{\mu}$ and $\phi_{\sigma^2}$ are the mean and variance estimation heads, respectively. Both maps retain the spatial structure of $\bm{E}$ and have $p$ channels. The mean map $\bm{E}_{d}$ represents the estimated degradation, while the variance map $\bm{E}_{g}$ provides an uncertainty cue for GER.

\noindent\textbf{Region-Aware Expert Routing.}~ 
To enable region-aware expert routing, we preserve the spatial dimensions of the degradation priors and equip each restoration block with its own DER and GER.
Let $\bm{H}^{(s)}$ be the input feature of the $s$-th block. The degradation-prior and restoration-feature branches of DER are denoted by $\mathcal{G}_{d,E}^{(s)}$ and $\mathcal{G}_{d,H}^{(s)}$, while those of GER are denoted by $\mathcal{G}_{g,E}^{(s)}$ and $\mathcal{G}_{g,H}^{(s)}$.
The routing logits are computed as,
\begin{align}
	\bm{Z}_{d}^{(s)}
	&=\mathcal{G}_{d,E}^{(s)}(\bm{E}_{d})
	+\mathcal{G}_{d,H}^{(s)}(\bm{H}^{(s)}),
	\label{equ:degradation_logits}\\
	\bm{Z}_{g}^{(s)}
	&=\mathcal{G}_{g,E}^{(s)}(\bm{E}_{g})
	+\mathcal{G}_{g,H}^{(s)}(\bm{H}^{(s)}),
	\label{equ:granularity_logits}
\end{align}
where $\bm{Z}_{d}^{(s)}$ and $\bm{Z}_{g}^{(s)}$ are the DER and GER logits, respectively. The two branches of each router produce logits on the same spatial grid for addition. At each spatial location, DER predicts $J_{L-1}$ logits for the finest-level groups, while GER predicts $L$ logits for levels $0,\ldots,L-1$.
Different locations can select experts according to their local degradation characteristics.
Conditioning both routers on $\bm{H}^{(s)}$ allows expert routing to adapt throughout the network.

At each spatial location, DER applies top-$1$ routing to $\bm{Z}_{d}^{(s)}$ to select a finest-level DR group, while GER applies top-$1$ routing to $\bm{Z}_{g}^{(s)}$ to select a hierarchy level. The selected finest-level group determines a path through the hierarchy, and the selected non-root level determines the expert on that path. Selecting level $0$ uses only the shared base block at that location.
Let $\bm{M}_{i,j}^{(s)}$ be the binary spatial mask assigned to expert $(i,j)$ by the joint decision of DER and GER. Each mask is aligned with its residual feature map and broadcast across channels. At each location, exactly one non-root mask is $1$ if GER selects a non-root level, and all non-root masks are $0$ if GER selects the root. The output of block $s$ is,
\begin{equation}
	\begin{aligned}
		\bm{H}^{(s+1)}={}&\mathcal{F}_{\mathrm{base}}^{(s)}(\bm{H}^{(s)})\\
		&+\sum_{i=1}^{L-1}\sum_{j=1}^{J_i}
		\bm{M}_{i,j}^{(s)}\odot
		\Delta\mathcal{F}_{i,j}^{(s)}(\bm{H}^{(s)}).
	\end{aligned}
	\label{equ:mask_fusion}
\end{equation}
For every block $s$ and spatial location $(u,v)$, $\sum_{i=1}^{L-1}\sum_{j=1}^{J_i} [\bm{M}_{i,j}^{(s)}]_{u,v}\in\{0,1\}$. The sum is $0$ when GER selects the root and $1$ otherwise, so each location uses the shared base block with at most one non-root expert residual.

As a result, our routing strategy preserves fine-grained specialization while allowing select coarser ancestor experts to alleviate the effect of degradation-estimate errors. This targets the assignment-related error in Eq.~(\ref{equ:hierarchical_scaling_model}) and helps preserve the restoration benefit of adding specialized experts.

\subsection{Loss Functions}
Let $\bm{U}$ denote the target map on $\Omega_d$, where each spatial location is assigned the finest-level cluster centroid of its corresponding source sample. For a CutMix input, its spatial mixing mask determines the source sample used at each target location.
We optimize the degradation distribution using the diagonal-Gaussian negative log-likelihood, omitting the
parameter-independent additive constant,
\begin{equation}
	\mathcal{L}_{dg}
	=\frac{1}{2|\Omega_d|}
	\sum_{r\in\Omega_d}\sum_{k=1}^{p}
	\left(
	\frac{\big(\bm{U}_{r,k}-\bm{E}_{d,r,k}\big)^2}
	{\bm{E}_{g,r,k}}
	+\log\bm{E}_{g,r,k}
	\right)
	,
	\label{equ:uncertainty_loss}
\end{equation}
where $r$ indexes locations in $\Omega_d$, $k$ indexes the $p$ DR channels, and $\log$ denotes the natural logarithm.
Thus, $\bm{E}_{g}$ provides a learned uncertainty cue for GER. The minimum value of $0.01$ in Eq.~(\ref{equ:degradation_distribution}) ensures a positive denominator and a well-defined logarithm.

In training restoration networks, we first train the shared base model and then train each LoRA expert on its corresponding DR group.
We then initialize the DERs using the finest-level cluster labels.
The routers and LoRA experts are then jointly fine-tuned with the CutMix pairs while the shared base model remains frozen.
We apply the load-balancing loss~\cite{sparse_gated_moe} $\mathcal{L}_{\mathrm{aux}}$ to GER to discourage collapse to a single granularity. The joint fine-tuning objective is,
\begin{equation}
	\mathcal{L}_{\mathrm{total}}
	=\ell_1+\alpha\mathcal{L}_{\mathrm{dg}}+\beta\mathcal{L}_{\mathrm{aux}},
	\label{equ:total_loss}
\end{equation}
where $\alpha$ and $\beta$ balance uncertainty learning and load balancing, respectively.

\section{Experiments}
\label{sec:experiments}

\begin{figure*}[!t]
	\centering
	\includegraphics[width=\textwidth]{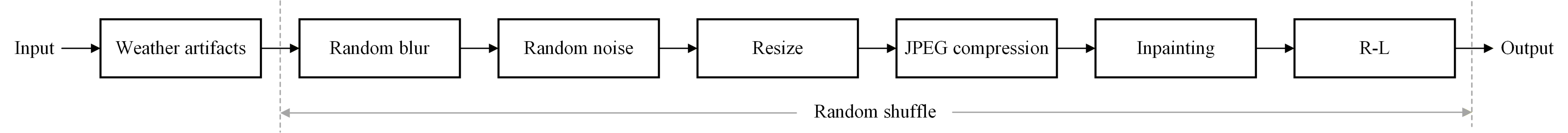}
	\caption{
    Illustration of the degradation pipeline. Weather artifacts are first introduced, after which the remaining degradation types are randomly selected, shuffled, and sequentially applied.
    }
	\label{fig:suppl_degradation_pipeline}
\end{figure*}

We first describe the datasets and evaluation settings in Section~\ref{sec:exp_setups}, followed by the implementation details in Section~\ref{sec:details}. 
We then compare \method~with state-of-the-art universal restoration methods in Section~\ref{sec:restor_perfor} and analyze the scaling behavior of universal restoration models in Section~\ref{sec:scalability}. 
We validate the proposed components through ablation studies in Section~\ref{sec:ablation_study}. Finally, we give the limitations and conclusions in Section~\ref{sec:conclusion}.

\subsection{Experimental Setup}
\label{sec:exp_setups}
We consider four universal image restoration settings:
the single-degradation setting, the triple-degradation setting, the penta-degradation setting, and the real-world evaluation setting.
The details are as follows.

\noindent\textbf{Single-Degradation Setting.}~  
Following PromptIR~\cite{PromptIR} and IDR~\cite{IDR}, we train on Rain100L~\cite{Rain200H} for deraining, RESIDE~\cite{reside} for dehazing, DFBW (DIV2K~\cite{div2k}, Flickr2K~\cite{flickr2k}, BSD~\cite{bsd68}, and WED~\cite{wed}) for Gaussian color image denoising, GoPro~\cite{GoPro} for motion deblurring, LOLv1~\cite{LOL} for low-light enhancement, Snow100K~\cite{Snow100K} for desnowing, and DFBW for JPEG compression artifact removal. We evaluate on Rain100L, SOTS~\cite{reside}, BSD68, GoPro, LOLv1, Snow100K, and LIVE1~\cite{live1}, respectively.

\noindent\textbf{Triple-Degradation Setting.}~
Following OneRestore~\cite{OneRestore}, we use CDD-11~\cite{OneRestore} for both training and evaluation. CDD-11 contains four first-order, five second-order, and two third-order degradation subsets.

\noindent\textbf{Penta-Degradation Setting.}~
We construct training data and an evaluation benchmark for a wider degradation space.

(1) \textit{Degradation space}: Starting from the Real-ESRGAN pipeline~\cite{wang2021realesrgan}, we introduce rain, snow, haze, mixed rain-haze, mixed snow-haze, inpainting, and Richardson-Lucy (R-L) artifacts~\cite{richardson1972bayesian, lucy1974iterative}. 
The degradation space also includes additional variants of existing degradations: noise includes Poisson, speckle, impulse, and spatially correlated noise, while blur includes jitter, glass, motion, lens, and zoom blur. To synthesize a training sample with $n_d$ degradation operations, we first select and apply one weather artifact, randomly sample the other $n_d-1$ degradations, shuffle them, and apply them sequentially. Fig.~\ref{fig:suppl_degradation_pipeline} illustrates the pipeline. 

(2) \textit{High-quality (HQ) training data}: We collect high-quality images from DIV2K~\cite{div2k}, Flickr2K~\cite{flickr2k}, DIV8K~\cite{gu2020div8k}, OutdoorSceneTraining~\cite{wang2018sftgan}, MIT-Adobe-5K~\cite{bychkovsky2011learning}, the first 5,000 images from FFHQ~\cite{ffhq}, and the first 500,000 images from SA-1B~\cite{Kirillov_2023_ICCV}.
We randomly crop five million $256\times256$ HQ patches and synthesize their low-quality (LQ) counterparts using the above pipeline, yielding five million paired training samples.

For evaluation, we develop CDD-50 using the clean images in Urban100~\cite{urban100}. 
CDD-50 comprises five degradation-order groups, ranging from one to five degradations, with ten degradation combinations per group generated by SEAL~\cite{seal}. The resulting 50 test settings are listed in Table~\ref{table:evaluation_configuration}.
\begin{table}[!t]
\caption{Degradation combinations in the five degradation-order groups of CDD-50. `RH' and `SH' denote mixed rain-haze and snow-haze degradations, respectively.}
\label{table:evaluation_configuration}
\centering
\setlength{\tabcolsep}{1mm}
\resizebox{\columnwidth}{!}{%
\begin{tabular}{c  c  c  c  c }
\toprule%

 1st-Order & 2nd-Order & 3rd-Order & 4th-Order & 5th-Order  \\

\midrule%
   Noise (N)      & I-RG   & B-I-J   & I-RL-J-RS  & RS-RL-B-N-RG   \\
   Blur (B)       & J-B    & RS-RG-J & RS-B-J-RG  & B-B-RL-N-J     \\
   JPEG (J)       & B-N    & B-N-RS  & RL-RL-RG-RG& SH-I-RL-RL-RL  \\
   Rain (R)       & RS-B   & RL-RL-J & RS-RG-RS-J & H-RL-B-RS-J    \\
   Snow (S)       & SH-RL  & B-RS-RL & RH-RS-RG-J & RH-RS-J-RL-RG  \\
   Haze (H)       & H-RG   & RS-RS-RL& N-B-J-RG   & J-J-I-RL-RG    \\
   Inpainting (I) & RS-J   & B-J-RG  & S-RL-J-J   & RS-RS-B-B-RL   \\
   R-L (RL)       & RG-RL  & RH-RG-J & N-RG-RS-J  & N-B-RS-J-RL    \\
   Resize (RS)    & N-J    & H-J-B   & B-J-RL-RG  & RL-I-B-I-RL    \\
   Ringing (RG)   & R-RL   & N-RL-RG & H-I-RL-J   & RG-RG-RG-I-RL  \\

\bottomrule%
\end{tabular}
}
\end{table}

Each test setting is evaluated with both in-distribution (\textit{in-dist.}) and out-of-distribution (\textit{out-of-dist.}) parameters, measuring restoration on the training parameter range and generalization to unseen parameter ranges, respectively.

\noindent\textbf{Real-World Setting.}~
We use SIDD~\cite{sidd}, RealBlur-J~\cite{realblur}, DPDD~\cite{dpdd}, RealRain~\cite{Wang_2019_CVPR}, RealSnow~\cite{WGWSNet}, WeatherStream~\cite{zhang2023weatherstream}, and FoundIR~\cite{foundir} to evaluate zero-shot generalization to real-world degradations.

\subsection{Implementation Details}
\label{sec:details}
We use Restormer~\cite{Restormer} as the restoration backbone.
We first train the DR extractor using the pixel-wise contrastive loss in Eq.~(\ref{equ:pixel_contrastive}) and construct the multi-granularity degradation set. 
We then train the shared base model, followed by the rank-$8$ LoRA experts~\cite{lorair} for their associated DR groups.
After expert training, we initialize the DERs using the finest-level cluster labels. 
Starting from this initialization, we jointly fine-tune the routers and LoRA experts while keeping the shared base model frozen.
We set $\alpha=0.1$ and $\beta=0.01$ in Eq.~(\ref{equ:total_loss}). 
Unless otherwise specified, we follow the training protocol of Restormer~\cite{Restormer}.
For the single-degradation setting, the three granularity levels contain $1$, $7$, and $19$ DR groups. For the triple-degradation setting, we use the three-level configuration $\{1,4,8\}$. For the penta-degradation setting, we construct the hierarchy as a binary tree. 
Training is performed using 64 NVIDIA V100 GPUs, each with 32\,GB of memory.
For the scaling experiments, we train each LoRA expert singly from scratch. Training $2.9$B model with $510$ LoRA experts takes approximately $23$ days.

\begin{table*}[!t]
\caption{Universal image restoration results in the single-degradation setting. CAR denotes compression artifact removal. PSNR (dB, $\uparrow$) and \colorbox{clblue!50}{SSIM ($\uparrow$)} are reported in RGB space. `$n$T' denotes a model jointly trained on $n$ tasks (3T: Derain, Dehaze, and Denoise; 5T: Derain, Dehaze, Denoise, Deblur, and Lowlight; 7T: all seven tasks). The weighted averages of benchmarks are reported. The \textbf{best} performance is highlighted.
}
\label{table:single_in_one_image_restoration_results}
\centering
\resizebox{\textwidth}{!}{%
\begin{tabular}{c  l  c c  c c  c c  c c  c c  c c  c c  c c c}
\toprule%

    \multirow{2}{*}{Type}
    & \multirow{2}{*}{Method}
    & \multicolumn{2}{c}{Derain}          
    & \multicolumn{2}{c}{Dehaze}        
    & \multicolumn{2}{c}{Denoise}      
    & \multicolumn{2}{c}{Deblur} 
    & \multicolumn{2}{c}{Low-light}  
    & \multicolumn{2}{c}{Desnow}           
    & \multicolumn{2}{c}{CAR}  
    & \multicolumn{3}{c}{Average}
    \\
    
    &
    & \multicolumn{2}{c}{\textcolor{gray}{Rain100L}} 
    & \multicolumn{2}{c}{\textcolor{gray}{SOTS}}
    & \multicolumn{2}{c}{\textcolor{gray}{BSD68}} 
    & \multicolumn{2}{c}{\textcolor{gray}{GoPro}} 
    & \multicolumn{2}{c}{\textcolor{gray}{LOLv1}} 
    & \multicolumn{2}{c}{\textcolor{gray}{Snow100K}}
    & \multicolumn{2}{c}{\textcolor{gray}{LIVE1}} 
    & \textcolor{gray}{3T} & \textcolor{gray}{5T} & \textcolor{gray}{7T}
    \\ 
\midrule%

    \multirow{5}{*}{3T}
    &
    AirNet~\cite{AirNet}             
    & 34.90 & \cc{.967} & 27.94 & \cc{.962} & 31.06 & \cc{.872} 
    & --    & \cc{--}   & --    & \cc{--}   & --    & \cc{--}   & -- & \cc{--} & 29.29 & -- & -- \\
    &
    PromptIR~\cite{PromptIR}        
    & 36.37 & \cc{.972} & 30.58 & \cc{.974} & 31.12 & \cc{.873}
    & --    & \cc{--}   & --    & \cc{--}   & --    & \cc{--}   & -- & \cc{--} & 31.50 & -- & -- \\
    &
    InstructIR~\cite{instructir}  
    & 37.98 & \cc{.978} & 30.22 & \cc{.959} & 31.32 & \cc{.875} 
    & --    & \cc{--}   & --    & \cc{--}   & --    & \cc{--}   & -- & \cc{--} & 31.49 & -- & -- \\
    &
    DaAIR~\cite{DaAIR} 
    & 37.10 & \cc{.978} & 32.30 & \cc{\textbf{.981}} & 31.06 & \cc{.868}
    & --    & \cc{--}   & --    & \cc{--}   & --    & \cc{--}   & -- & \cc{--} & 32.51 & -- & -- \\
    &
    MoCE-IR~\cite{moce} 
    & 38.57 & \cc{.984} & 31.34 & \cc{.979} & 31.24 & \cc{.873}
    & --    & \cc{--}   & --    & \cc{--}   & --    & \cc{--}   & -- & \cc{--} & 32.73  & -- & -- \\

\midrule%

    \multirow{5}{*}{5T}
    &
    AirNet~\cite{AirNet} 
    & 32.98 & \cc{.951} & 21.04 & \cc{.884} & 30.91 & \cc{.882} & 24.35 & \cc{.781}
    & 18.18 & \cc{.735} & --    & \cc{--}   & --    & \cc{--}   & 23.82 & 24.10 & --\\
    &
    InstructIR~\cite{instructir} 
    & 36.84 & \cc{.973} & 27.10 & \cc{.956} & 31.40 & \cc{.887} & 29.40 & \cc{.886}
    & 23.00 & \cc{.836} & --    & \cc{--}   & --    & \cc{--}   & 28.99 & 29.19 & --\\
    &
    DaAIR~\cite{DaAIR}
    & 36.28 & \cc{.975} & 31.97 & \cc{.980} & 31.07 & \cc{.878} & 29.51 & \cc{.890}
    & 22.38 & \cc{.825} & --    & \cc{--}   & --    & \cc{--}   & 32.20 & 30.24 & -- \\ 
    &
    PromptIR-DCPT~\cite{dcpt}
    & 37.32 & \cc{.978} & 30.72 & \cc{.977} & 31.32 & \cc{.885} & 28.84 & \cc{.877}
    & 23.35 & \cc{.840} & --    & \cc{--}   & --    & \cc{--}   & 31.46 & 30.31 & -- \\
    &
    MoCE-IR~\cite{moce}
    & 38.04 & \cc{.982} & 30.48 & \cc{.974} & 31.34 & \cc{.887} & 30.05 & \cc{.899}
    & 23.00 & \cc{\textbf{.852}} & --    & \cc{--}   & --    & \cc{--}   & 31.39 & 30.58 & -- \\
    
\midrule%

    \multirow{2}{*}{7T} 
    &
    UniRestorer~\cite{lin2026unirestorer}%
    & 40.22 & \cc{\textbf{.991}} & 34.94 & \cc{.971} & 31.39 & \cc{\textbf{.937}} & 30.86 & \cc{.933}
    & 24.42 & \cc{.827} & 30.53 & \cc{.925} & 30.31 & \cc{.922} & 35.51 & 32.36 & 31.81 \\
    &
    \method%
    & \textbf{40.35} & \cc{\textbf{.991}} & \textbf{35.51} & \cc{.980} & \textbf{31.48} & \cc{\textbf{.937}} & \textbf{31.59} & \cc{\textbf{.950}}
    & \textbf{24.64} & \cc{.831} & \textbf{30.66} & \cc{\textbf{.928}} & \textbf{31.48} & \cc{\textbf{.947}} & \textbf{35.78} & \textbf{32.71} & \textbf{32.24} \\

\bottomrule%
\end{tabular}
}
\end{table*}

\begin{table*}[!t]
\caption{Universal image restoration results on CDD-11. L, H, R, and S denote low-light, haze, rain, and snow, respectively. PSNR (dB, $\uparrow$) and \colorbox{clblue!50}{SSIM ($\uparrow$)} are reported in RGB space. 
The \textbf{best} performance is highlighted.}
\label{table:cdd11_results}
\centering
\resizebox{\textwidth}{!}{%
\begin{tabular}{l | cc  cc  cc  cc  | cc  cc cc cc cc | cc cc  | c}
\toprule%

    \multirow{2}{*}{Method}

    & \multicolumn{8}{c|}{1st-Order}
    & \multicolumn{10}{c|}{2nd-Order}
    & \multicolumn{4}{c|}{3rd-Order}
    & \multirow{2}{*}{Average}
    \\
    
    & \multicolumn{2}{c}{\textcolor{gray}{L}}
    & \multicolumn{2}{c}{\textcolor{gray}{H}}
    & \multicolumn{2}{c}{\textcolor{gray}{R}}
    & \multicolumn{2}{c|}{\textcolor{gray}{S}}
    & \multicolumn{2}{c}{\textcolor{gray}{L-H}}
    & \multicolumn{2}{c}{\textcolor{gray}{L-R}}
    & \multicolumn{2}{c}{\textcolor{gray}{L-S}}
    & \multicolumn{2}{c}{\textcolor{gray}{H-R}}
    & \multicolumn{2}{c|}{\textcolor{gray}{H-S}}
    & \multicolumn{2}{c}{\textcolor{gray}{L-H-R}}
    & \multicolumn{2}{c|}{\textcolor{gray}{L-H-S}}
    & 
    \\
        
\midrule%
AirNet~\cite{AirNet} 
    & 24.83&\cc{.778} & 24.21&\cc{.951} & 26.55&\cc{.891} & 26.79&\cc{.919}
    & 23.23&\cc{.779} & 22.82&\cc{.710} & 23.29&\cc{.723} & 22.21&\cc{.868} & 23.29&\cc{.901}
    & 21.80&\cc{.708} & 22.24&\cc{.725} & 23.75 \\
PromptIR~\cite{PromptIR} 
    & 26.32&\cc{.805} & 26.10&\cc{.969} & 31.56&\cc{.946} & 31.53&\cc{.960} 
    & 24.49&\cc{.789} & 25.05&\cc{.771} & 24.51&\cc{.761} & 24.54&\cc{.924} & 23.70&\cc{.925} 
    & 23.74&\cc{.752} & 23.33&\cc{.747} & 25.90 \\
OneRestore~\cite{OneRestore}
    & 26.48&\cc{.826} & 32.52&\cc{.990} & 33.40&\cc{.964} & 34.31&\cc{.973}  
    & 25.79&\cc{.822} & 25.58&\cc{.799} & 25.19&\cc{.789} & 29.99&\cc{.957} & 30.21&\cc{.964} 
    & 24.78&\cc{.788} & 24.90&\cc{.791} & 28.47 \\
MoCE-IR~\cite{moce}
    & 27.26&\cc{.824} & 32.66&\cc{.990} & 34.31&\cc{.970} & 35.91&\cc{.980}
    & 26.24&\cc{.817} & 26.25&\cc{.800} & 26.04&\cc{.793} & 29.93&\cc{\textbf{.964}} & 30.19&\cc{.970}
    & 25.41&\cc{.789} & 25.39&\cc{.790} & 29.05 \\
    
\midrule%

UniRestorer~\cite{lin2026unirestorer}%
    & 27.71 & \cc{.900} & 34.53 & \cc{.991} & 36.07 & \cc{.972} & 36.84 & \cc{\textbf{.983}}
    & 26.88 & \cc{.893} & 26.93 & \cc{.894} & 26.73 & \cc{.892} & 32.02 & \cc{.936}
    & 30.82 & \cc{.976} & 25.78 & \cc{\textbf{.884}} & 25.87 & \cc{.885} & 30.02\\

\method~%
    & \textbf{27.93} & \cc{\textbf{.901}} & \textbf{34.98} & \cc{\textbf{.992}} & \textbf{36.28} & \cc{\textbf{.975}} & \textbf{36.89} & \cc{\textbf{.983}}
    & \textbf{27.23} & \cc{\textbf{.896}} & \textbf{27.22} & \cc{\textbf{.898}} & \textbf{27.05} & \cc{\textbf{.893}} & \textbf{32.66} & \cc{.940}
    & \textbf{31.19} & \cc{\textbf{.980}} & \textbf{25.92} & \cc{.883} & \textbf{26.11} & \cc{\textbf{.887}} & \textbf{30.31} \\

\bottomrule%
\end{tabular}
}
\end{table*}

\begin{table*}[!t]
\caption{Universal image restoration results on CDD-50 under \textit{in-dist.} and \textit{out-of-dist.} degradation parameters. PSNR (dB, $\uparrow$) and \colorbox{clblue!50}{SSIM ($\uparrow$)} are reported in RGB space. The \textbf{best} performance is highlighted.}
\label{table:ourset_results}
\centering
\resizebox{\textwidth}{!}{%
\begin{tabular}{l | cc  cc  cc  cc  cc  cc | cc cc cc cc cc cc }
\toprule%

    \multirow{2}{*}{Method}
    & \multicolumn{12}{c|}{In-dist. \,\, Degradation}
    & \multicolumn{12}{c}{Out-of-dist. \,\, Degradation}
    \\

    & \multicolumn{2}{c}{\textcolor{gray}{1st-Order}}
    & \multicolumn{2}{c}{\textcolor{gray}{2nd-Order}}
    & \multicolumn{2}{c}{\textcolor{gray}{3rd-Order}}
    & \multicolumn{2}{c}{\textcolor{gray}{4th-Order}}
    & \multicolumn{2}{c}{\textcolor{gray}{5th-Order}}
    & \multicolumn{2}{c|}{\textcolor{gray}{Average}}

    & \multicolumn{2}{c}{\textcolor{gray}{1st-Order}}
    & \multicolumn{2}{c}{\textcolor{gray}{2nd-Order}}
    & \multicolumn{2}{c}{\textcolor{gray}{3rd-Order}}
    & \multicolumn{2}{c}{\textcolor{gray}{4th-Order}}
    & \multicolumn{2}{c}{\textcolor{gray}{5th-Order}}
    & \multicolumn{2}{c}{\textcolor{gray}{Average}}
    \\
    
\midrule
AirNet~\cite{AirNet}
& 28.25 & \cc{.912} & 23.10 & \cc{.808} & 20.50 & \cc{.758} & 20.20 & \cc{.760} & 19.05 & \cc{.714} & 22.22 & \cc{.790} 
& 25.60 & \cc{.868} & 20.35 & \cc{.755} & 18.80 & \cc{.710} & 18.45 & \cc{.690} & 16.70 & \cc{.655} & 19.98 & \cc{.736} \\
PromptIR~\cite{PromptIR} 
& 30.78 & \cc{.943} & 25.24 & \cc{.848} & 22.27 & \cc{.801} & 22.11 & \cc{.806} & 20.92 & \cc{.761} & 24.26 & \cc{.832} 
& 27.92 & \cc{.902} & 22.45 & \cc{.795} & 20.66 & \cc{.751} & 19.92 & \cc{.724} & 18.01 & \cc{.689} & 21.79 & \cc{.772} \\
OneRestore~\cite{OneRestore} 
& 28.28 & \cc{.926} & 24.02 & \cc{.846} & 22.54 & \cc{.793} & 22.15 & \cc{.792} & 20.72 & \cc{.735} & 23.54 & \cc{.818} 
& 26.08 & \cc{.897} & 21.42 & \cc{.788} & 20.19 & \cc{.731} & 20.08 & \cc{.727} & 17.85 & \cc{.642} & 21.12 & \cc{.757} \\
MoCE-IR~\cite{moce} 
& 31.15 & \cc{.948} & 26.36 & \cc{.882} & 23.52 & \cc{.823} & 23.32 & \cc{.821} & 22.07 & \cc{.775} & 25.28 & \cc{.850} 
& 28.56 & \cc{.923} & 23.66 & \cc{.830} & 21.24 & \cc{.776} & 20.57 & \cc{.758} & 18.60 & \cc{\textbf{.721}} & 22.53 & \cc{.802} \\

\midrule

UniRestorer~\cite{lin2026unirestorer} 
& 32.58 & \cc{.959} & 26.98 & \cc{.901} & 24.29 & \cc{.842} & 23.81 & \cc{.837} & 22.53 & \cc{.792} & 26.04 & \cc{.866} 
& 29.72 & \cc{.935} & 23.91 & \cc{.845} & 21.57 & \cc{.781} & 21.13 & \cc{.769} & 19.12 & \cc{.713} & 23.09 & \cc{.809} \\
\method
& \textbf{33.54} & \cc{\textbf{.967}} & \textbf{27.40} & \cc{\textbf{.913}} & \textbf{24.80} & \cc{\textbf{.855}} & \textbf{24.14} & \cc{\textbf{.848}} & \textbf{22.84} & \cc{\textbf{.804}} & \textbf{26.54} & \cc{\textbf{.877}} 
& \textbf{30.49} & \cc{\textbf{.943}} & \textbf{24.07} & \cc{\textbf{.855}} & \textbf{21.79} & \cc{\textbf{.785}} & \textbf{21.51} & \cc{\textbf{.777}} & \textbf{19.46} & \cc{.708} & \textbf{23.46} & \cc{\textbf{.813}} \\

\bottomrule%
\end{tabular}
}
\end{table*}

\begin{table}[!t]
\caption{
Zero-shot generalization comparison on real-world datasets. PSNR (dB, $\uparrow$) is reported in RGB space. All methods are trained in the single-degradation setting and evaluated without additional training. The \textbf{best} performance is highlighted.}
\label{table:zero_shot_generalization}
\centering
\setlength{\tabcolsep}{2mm}
\resizebox{\columnwidth}{!}{%
\begin{tabular}{ l  c  c c  c c c c}
\toprule%

    \multirow{2}{*}{Method} 
    & SIDD%
    & RealBlur-J%
    & DPDD%
    & RealRain%
    & RealSnow%
    & WeatherStream
    & FoundIR%
    \\

    & \cite{sidd}
    & \cite{realblur}
    & \cite{dpdd}
    & \cite{Wang_2019_CVPR}
    & \cite{WGWSNet}
    & \cite{zhang2023weatherstream}
    & \cite{foundir}
    \\
    
\midrule%
    
    AirNet~\cite{AirNet}          & 23.65 & 24.92 & 20.17 & 22.56 & 23.63 & 21.68  & 20.56  \\ 
    PromptIR~\cite{PromptIR}      & 24.32 & 25.47 & 21.76 & 24.12 & 24.94 & 22.09 & 21.54 \\        
    OneRestore~\cite{OneRestore}  & 11.46 & 18.79 & 20.96 & 22.73 & 20.72 & 19.49 & 22.94\\   
    MoCE-IR~\cite{moce}           & 15.96 & 14.29 & 15.61 & 28.10 & 24.63 & 20.54 & 22.63 \\        
    
\midrule%

    UniRestorer~\cite{lin2026unirestorer}    & 24.70 & 25.55 & 23.60 & 29.85 & 27.73 & 22.25 & 23.80 \\   
    \method                                 & \textbf{24.85} & \textbf{26.25} & \textbf{23.90} & \textbf{29.95} & \textbf{27.80} & \textbf{22.55} & \textbf{24.05} \\

\bottomrule%
\end{tabular}
}
\end{table}

\begin{table}[!t]
\caption{Computational cost comparison with state-of-the-art universal methods in the single-degradation setting.} 
\label{table:overhead_comparison}
\centering
\setlength{\tabcolsep}{3.5mm}
\resizebox{\columnwidth}{!}{%
\begin{tabular}{ l c c c }
\toprule%
Method               & \#Params. (M) & \#FLOPs (G) & Latency (s) \\
\midrule%
AirNet~\cite{AirNet}          & 9     & 2414   & 0.548 \\
PromptIR~\cite{PromptIR}      & 36    & 1266   & 0.355 \\
InstructIR~\cite{instructir}  & 16    & 130    & 0.055 \\
PromptIR-DCPT~\cite{dcpt}     & 36    & 1266   & 0.355  \\
MoCE-IR~\cite{moce}           & 25    & 716    & 0.347  \\

\midrule
UniRestorer~\cite{lin2026unirestorer}%
& 171 & 1156 & 0.484  \\
\method%
& 176 & 1255 & 0.515  \\

\bottomrule%
\end{tabular}
}
\end{table}

\subsection{Method Comparison and Results}
\label{sec:restor_perfor}
\begin{figure*}[!t]
	\centering
	\includegraphics[width=\textwidth]{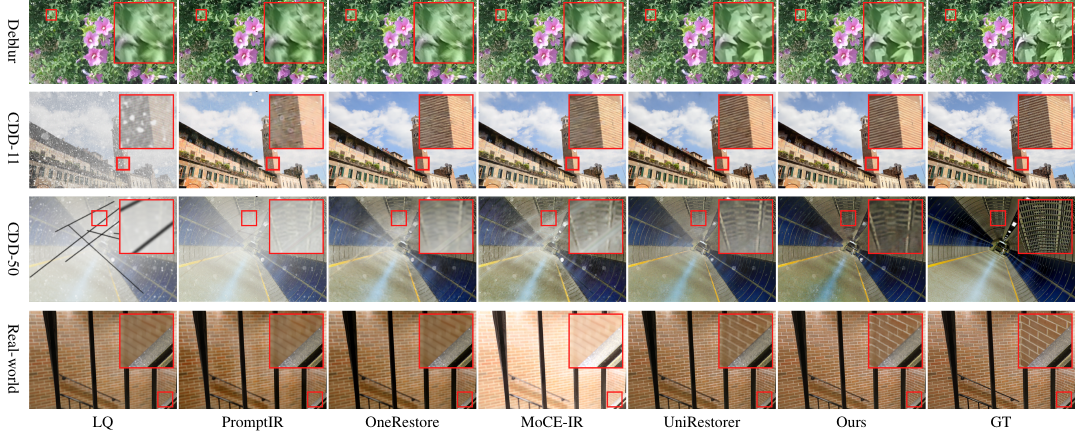}
	\caption{Visual comparison across the single-degradation (GoPro deblurring), triple-degradation (CDD-11), penta-degradation (CDD-50), and real-world settings. \method~removes degradations and recovers textures and colors, as highlighted by the \textcolor{red}{red boxes} and enlarged regions.}
    \label{fig:visual_comparison}
\end{figure*}

\noindent\textbf{Performance Comparison.}~ 
Tables~\ref{table:single_in_one_image_restoration_results} to \ref{table:zero_shot_generalization} compare the methods on single, composite, and real-world degradations. Our method achieves the state-of-the-art performance among the compared universal methods in every evaluated task or degradation-order group.
Compared to UniRestorer, \method~raises the seven-task average from $31.81$ to $32.24$ dB, reaches $30.31$ dB on CDD-11 ($+0.29$ dB), and improves CDD-50 from $26.04$/$23.09$ to $26.54$/$23.46$ dB on the \textit{in-dist.}/\textit{out-of-dist.} settings. 
These consistent gains demonstrate the effectiveness of the proposed method across increasingly complex degradation spaces.
Fig.~\ref{fig:visual_comparison} compares restoration results across the four experimental settings, showing that our method can recover fine structure and texture. 

\noindent\textbf{Computational Cost Comparison.}~ 
Table~\ref{table:overhead_comparison} reports parameters, FLOPs, and latency for a $512\times512\times3$ input on a single NVIDIA A6000 GPU, averaged over 100 inference runs with randomly sampled degradations from the single-degradation setting.
Compared to UniRestorer, \method~increases the parameter count from $171$M to $176$M ($2.9\%$), FLOPs from $1156$G to $1255$G ($8.6\%$), and latency from $0.484$ s to $0.515$ s ($6.4\%$). Thus, the proposed routing design introduces slight inference overhead while improving restoration performance.
Moreover, top-$1$ routing activates at most one non-root LoRA expert per spatial location in each restoration block, keeping the active expert computation approximately fixed as total capacity grows. This allows \method~to scale from $176$M to $2.9$B parameters while maintaining approximately unchanged FLOPs and inference latency.

\begin{table*}[!t]
\caption{Performance of hierarchically scaled \method~variants. PSNR (dB, $\uparrow$) and \colorbox{clblue!50}{SSIM ($\uparrow$)} are reported in RGB space. All \method~variants are trained on the complete CDD-50 degradation space. Each degradation-order row uses its own $238$M baseline trained only on the corresponding order, and all improvements are computed relative to the baseline in the same row. In the real-world setting, the $237.8$M baseline is trained on the complete CDD-50 degradation space containing all five orders.}
\label{table:scaling_up_ours_average}
\centering
\setlength{\tabcolsep}{4mm}
\resizebox{\textwidth}{!}{%
\begin{tabular}{ c c | c c | cc cc cc cc cc cc}
\toprule

Evaluation
& \multirow{2}{*}{\makecell[c]{Degradation Order\\/ Test Set}}
& \multicolumn{2}{c|}{Baseline}  
& \multicolumn{2}{c}{\method}
& \multicolumn{2}{c}{\method}
& \multicolumn{2}{c}{\method}
& \multicolumn{2}{c}{\method}
& \multicolumn{2}{c}{\method}
& \multicolumn{2}{c}{\method}
\\

Setting
&
& \multicolumn{2}{c|}{\textcolor{gray}{238M}}
& \multicolumn{2}{c}{\textcolor{gray}{154M}} 
& \multicolumn{2}{c}{\textcolor{gray}{244M}}
& \multicolumn{2}{c}{\textcolor{gray}{423M}}
& \multicolumn{2}{c}{\textcolor{gray}{781M}}
& \multicolumn{2}{c}{\textcolor{gray}{1.5B}}
& \multicolumn{2}{c}{\textcolor{gray}{2.9B}}
\\
\midrule

\multirow{5}{*}{\makecell[c]{In-dist. \\ Synthetic \\ Degradation \\ in CDD-50}}
  & 1st-Order & 33.39 & \cc{.965} & +0.15 & \cc{+.002} & +0.36 & \cc{+.003} & +0.56 & \cc{+.003} & +0.81 & \cc{+.004} & +0.90 & \cc{+.005} & +0.94 & \cc{+.006} \\
  & 2nd-Order & 27.35 & \cc{.912} & +0.05 & \cc{+.001} & +0.15 & \cc{+.002} & +0.31 & \cc{+.003} & +0.46 & \cc{+.005} & +0.55 & \cc{+.005} & +0.58 & \cc{+.006} \\
  & 3rd-Order & 24.76 & \cc{.854} & +0.04 & \cc{+.001} & +0.10 & \cc{+.002} & +0.21 & \cc{+.003} & +0.31 & \cc{+.005} & +0.40 & \cc{+.007} & +0.44 & \cc{+.008} \\
  & 4th-Order & 24.01 & \cc{.845} & +0.13 & \cc{+.003} & +0.19 & \cc{+.003} & +0.30 & \cc{+.005} & +0.44 & \cc{+.017} & +0.51 & \cc{+.018} & +0.55 & \cc{+.019} \\
  & 5th-Order & 22.78 & \cc{.803} & +0.06 & \cc{+.001} & +0.14 & \cc{+.003} & +0.23 & \cc{+.004} & +0.36 & \cc{+.006} & +0.43 & \cc{+.008} & +0.46 & \cc{+.009} \\
    
\midrule

\multirow{5}{*}{\makecell[c]{Out-of-dist. \\ Synthetic \\ Degradation \\ in CDD-50}}
  & 1st-Order & 30.35 & \cc{.942} & +0.14 & \cc{+.001} & +0.31 & \cc{+.002} & +0.50 & \cc{+.002} & +0.71 & \cc{+.003} & +0.81 & \cc{+.004} & +0.85 & \cc{+.005} \\
  & 2nd-Order & 23.90 & \cc{.853} & +0.17 & \cc{+.002} & +0.25 & \cc{+.003} & +0.37 & \cc{+.005} & +0.49 & \cc{+.007} & +0.58 & \cc{+.009} & +0.62 & \cc{+.010} \\
  & 3rd-Order & 21.62 & \cc{.782} & +0.17 & \cc{+.003} & +0.23 & \cc{+.004} & +0.31 & \cc{+.007} & +0.41 & \cc{+.009} & +0.50 & \cc{+.012} & +0.54 & \cc{+.013} \\
  & 4th-Order & 21.46 & \cc{.775} & +0.05 & \cc{+.002} & +0.13 & \cc{+.003} & +0.23 & \cc{+.006} & +0.37 & \cc{+.008} & +0.44 & \cc{+.011} & +0.48 & \cc{+.012} \\
  & 5th-Order & 19.23 & \cc{.703} & +0.23 & \cc{+.005} & +0.30 & \cc{+.007} & +0.39 & \cc{+.009} & +0.51 & \cc{+.012} & +0.57 & \cc{+.014} & +0.60 & \cc{+.015} \\

\midrule
\multirow{7}{*}{\makecell[c]{Real-world \\ Degradation}}
  & SIDD~\cite{sidd}                            
  & 23.94 & \cc{.508} & +6.48 & \cc{+.232} & +8.65 & \cc{+.301} & +11.04 & \cc{+.353} & +12.63 & \cc{+.387} & +13.74 & \cc{+.411} & +14.78 & \cc{+.434} \\
  & RealBlur-J~\cite{realblur}                  
  & 26.42 & \cc{.850} & +1.17 & \cc{+.005} & +2.38 & \cc{+.011} & +3.16 & \cc{+.028} & +3.68 & \cc{+.039} & +4.05 & \cc{+.047} & +4.39 & \cc{+.055} \\
  & DPDD~\cite{dpdd}                            
  & 23.86 & \cc{.731} & +0.22 & \cc{+.018} & +0.60 & \cc{+.033} & +0.78 & \cc{+.040} & +0.89 & \cc{+.045} & +0.97 & \cc{+.048} & +1.05 & \cc{+.051} \\
  & RealRain~\cite{Wang_2019_CVPR}              
  & 28.89 & \cc{.894} & +1.36 & \cc{+.016} & +1.85 & \cc{+.024} & +2.10 & \cc{+.028} & +2.26 & \cc{+.032} & +2.37 & \cc{+.034} & +2.48 & \cc{+.036} \\
  & RealSnow~\cite{WGWSNet}                     
  & 27.40 & \cc{.884} & +0.40 & \cc{-.004} & +1.00 & \cc{-.002} & +1.19 & \cc{+.003} & +1.31 & \cc{+.006} & +1.40 & \cc{+.009} & +1.48 & \cc{+.011} \\
  & WeatherStream~\cite{zhang2023weatherstream}                     
  & 21.22 & \cc{.726} & +1.53 & \cc{+.064} & +1.75 & \cc{+.071} & +1.88 & \cc{+.075} & +1.96 & \cc{+.078} & +2.02 & \cc{+.079} & +2.08 & \cc{+.081} \\
  & FoundIR~\cite{foundir}
  & 25.88 & \cc{.797} & +0.34 & \cc{+.011} & +0.67 & \cc{+.032} & +1.18 & \cc{+.040} & +1.65 & \cc{+.048} & +1.98 & \cc{+.054} & +2.29 & \cc{+.060} \\
\bottomrule
\end{tabular}
}
\end{table*}

\begin{table*}[!t]
\caption{Zero-, one-, and five-shot comparison with state-of-the-art specialist models on real-world benchmarks. Specialist results are taken from the corresponding papers or trained from scratch. PSNR (dB, $\uparrow$) and \colorbox{clblue!50}{SSIM ($\uparrow$)} are reported in RGB space. We repeat each few-shot experiment five times and report $\operatorname{mean}\pm\operatorname{std}$. The \textbf{best} performance is highlighted.}
\label{table:scaling_up_compare_to_specialists}
\centering
\setlength{\tabcolsep}{2.5mm}
\resizebox{\textwidth}{!}{%
\begin{tabular}{l l l cc cc cc cc}
\toprule%

\multirow{2}{*}{Test Sets} 
& \multirow{2}{*}{Task}
& \multirow{2}{*}{Specialist Methods}
& \multicolumn{2}{c}{Performance}
& \multicolumn{2}{c}{\method}
& \multicolumn{2}{c}{\method}
& \multicolumn{2}{c}{\method}
\\

& 
&
& \multicolumn{2}{c}{of Specialists} 
& \multicolumn{2}{c}{\textcolor{gray}{(2.9B, zero-shot)}}
& \multicolumn{2}{c}{\textcolor{gray}{(2.9B, 1-shot)}}
& \multicolumn{2}{c}{\textcolor{gray}{(2.9B, 5-shot)}}

\\
\midrule%

SIDD~\cite{sidd} & Denoising & NAFNet~\cite{NAFNet}                       
 & \textbf{40.30} & \cc{\textbf{.962}} & 38.72 & \cc{.942} & 38.83 $\pm$ .21 & \cc{.945} $\pm$ .009 & 39.26 $\pm$ .13 & \cc{.957} $\pm$ .004 \\ 
RealBlur-J~\cite{realblur} & Motion Deblurring & X-Restormer~\cite{xrestormer}
 & 36.27 & \cc{\textbf{.958}} & 30.81 & \cc{.905} & 36.20 $\pm$ .11 & \cc{.952} $\pm$ .005 & \textbf{36.28} $\pm$ .05 & \cc{\textbf{.958}} $\pm$ .002 \\
DPDD~\cite{dpdd} & Defocus Deblurring & Restormer~\cite{Restormer}
 & 25.98 & \cc{.811} & 24.91 & \cc{.782} & 25.81 $\pm$ .28 & \cc{.804} $\pm$ .015 & \textbf{26.28} $\pm$ .14 & \cc{\textbf{.815}} $\pm$ .006 \\
RealRain~\cite{Wang_2019_CVPR} & Deraining & Restormer~\cite{Restormer} & 43.93 & \cc{\textbf{.993}} & 31.37 & \cc{.930} & 41.95 $\pm$ .37 & \cc{.990} $\pm$ .003 & \textbf{44.06} $\pm$ .16 & \cc{.991} $\pm$ .002 \\ 
RealSnow~\cite{WGWSNet} & Snow Removal & Restormer~\cite{Restormer} & 30.94 & \cc{.913} & 28.88 & \cc{.895} & 31.58 $\pm$ .42 & \cc{.946} $\pm$ .011 & \textbf{33.28} $\pm$ .19 & \cc{\textbf{.969}} $\pm$ .004 \\ 
WeatherStream~\cite{zhang2023weatherstream} & Deweathering & Restormer (Imper.)~\cite{liu2024learning} & 23.15 & \cc{.810} & 23.30 & \cc{.807} & 23.35 $\pm$ .24 & \cc{.815} $\pm$ .009 & \textbf{23.72} $\pm$ .15 & \cc{\textbf{.825}} $\pm$ .006 \\
\bottomrule%
\end{tabular}
}
\end{table*}

\begin{table*}[!t]
\caption{One- and five-shot performance on FoundIR~\cite{foundir}. PSNR (dB, $\uparrow$) and \colorbox{clblue!50}{SSIM ($\uparrow$)} are reported in RGB space. We repeat each experiment five times and report $\operatorname{mean}\pm\operatorname{std}$. The \textbf{best} performance is highlighted.}
\label{table:foundir}
\centering
\resizebox{\textwidth}{!}{%
 \begin{tabular}{l  c  c  c  c  c  c c c c c c}
\toprule%

    Method
    & Blur %
    & Noise %
    & JPEG %
    & Haze %
    & Rain %
    & Raindrop %
    & B+N %
    & B+J %
    & N+J %
    & R+H %
    & B+N+J %
    \\
            
\midrule%
\multirow{2}{*}{FoundIR~\cite{foundir}} 
    & 25.93  & 38.98  & 34.03  & 22.19
    & 34.29  & 29.74  & 23.87  & \textbf{29.58}  & 40.43
    & 21.80  & \textbf{27.98}  \\
    & \cc{.791} & \cc{.964} & \cc{.942} & \cc{.849}
    & \cc{.943} & \cc{.916} & \cc{.773} & \cc{\textbf{.854}} & \cc{.979}
    & \cc{.758} & \cc{\textbf{.745}} \\

\midrule%

\method
& 25.80 $\pm$ .22 & 40.10 $\pm$ .11  & 34.80 $\pm$ .08
& 21.50 $\pm$ .37 & 34.80 $\pm$ .26  & 32.80 $\pm$ .31 
& 24.90 $\pm$ .28 & 26.80 $\pm$ .34  & 40.20 $\pm$ .14 
& 23.00 $\pm$ .36 & 27.10 $\pm$ .32  \\
\textcolor{gray}{(2.9B, 1-shot)}
& \cc{.775 $\pm$ .006} & \cc{.970 $\pm$ .003} & \cc{.975 $\pm$ .002}
& \cc{.852 $\pm$ .008} & \cc{.949 $\pm$ .005} & \cc{.945 $\pm$ .006} 
& \cc{.832 $\pm$ .007} & \cc{.818 $\pm$ .007} & \cc{.973 $\pm$ .002}
& \cc{.812 $\pm$ .010} & \cc{.725 $\pm$ .009} \\
    
\method
& \textbf{27.62} $\pm$ .14 & \textbf{41.89} $\pm$ .04 & \textbf{35.23} $\pm$ .02 
& \textbf{23.34} $\pm$ .23 & \textbf{37.10} $\pm$ .15 & \textbf{37.22} $\pm$ .18 
& \textbf{27.02} $\pm$ .21 & 25.94 $\pm$ .26 & \textbf{41.01} $\pm$ .05
& \textbf{26.81} $\pm$ .29 & 27.30 $\pm$ .12 \\
\textcolor{gray}{(2.9B, 5-shot)}
& \cc{\textbf{.881} $\pm$ .004} & \cc{\textbf{.985} $\pm$ .002} & \cc{\textbf{.976} $\pm$ .002}
& \cc{\textbf{.894} $\pm$ .006} & \cc{\textbf{.969} $\pm$ .003} & \cc{\textbf{.979} $\pm$ .003} 
& \cc{\textbf{.864} $\pm$ .005} & \cc{.803 $\pm$ .006} & \cc{\textbf{.985} $\pm$ .002}
& \cc{\textbf{.891} $\pm$ .006} & \cc{.742 $\pm$ .007}  \\
    
\bottomrule%
\end{tabular}
}
\end{table*}

\subsection{Scaling Universal Restoration Models}
\label{sec:scalability}

We compare three scaling settings on CDD-50 and FoundIR: dense model scaling enlarges the depth and width of Restormer (five variants, $26$M to $702$M); vanilla MoE scaling increases the number of finest-level experts (six variants, $120$M to $1.5$B); and our hierarchy model scaling retains experts at multiple granularities. 
In addition, Table~\ref{table:scaling_up_ours_average} includes baseline models with $238$M parameters. On CDD-50, separate baselines are trained on each degradation order.

\noindent\textbf{Hierarchy Model Scaling Strategy.}~
In our method, we implement the group-specific parameters as LoRA experts, use the $76$M Restormer as the shared base model, and increase the number of non-root binary hierarchy levels from three to eight, corresponding to $14$, $30$, $62$, $126$, $254$, and $510$ LoRA experts. Each added level refines the degradation partition and introduces parameters for the resulting fine-grained groups.
This produces six \method~variants with $154$M, $244$M, $423$M, $781$M, $1.5$B, and $2.9$B parameters.

Table~\ref{table:scaling_up_ours_average} and Fig.~\ref{fig:scaling_up_params_psnr_ours} show consistent gains as the hierarchical model grows from $154$M to $2.9$B parameters: average PSNR increases from $26.54$ to $27.05$ dB on \textit{in-dist.} CDD-50, from $23.46$ to $23.93$ dB on \textit{out-of-dist.} CDD-50, and from $26.22$ to $28.17$ dB on FoundIR.
We fit the reported average PSNR of each strategy using the empirical form in Eq.~(\ref{equ:predicted_log_scaling}), with $P_{\mathrm{total}}$ measured in millions and $a_\kappa^{\mathrm{fit}},b_\kappa^{\mathrm{fit}}$ fitted over the evaluated range. The fits of our hierarchy model scaling yield approximate $R^2$ values of $0.95$, $0.96$, and $0.98$, respectively.
Both MoE strategies show approximately log-linear trends over the evaluated range, whereas dense model scaling is non-monotonic. Their fitted slopes $b_\kappa^{\mathrm{fit}}$ are similar on \textit{in-dist.} CDD-50, but ours is steeper on \textit{out-of-dist.} CDD-50 ($0.386$ versus $0.133$) and FoundIR ($1.555$ versus $0.808$). Thus, our hierarchy model scaling provides larger gains from additional capacity under \textit{out-of-dist.} and on real-world degradations, with its advantage widening over the shared parameter range.
Our models also achieve higher PSNR than the $238$M baseline models in Table~\ref{table:scaling_up_ours_average} and lie above the vanilla MoE scaling curves at comparable total parameter counts in Fig.~\ref{fig:scaling_up_params_psnr_ours}.

\noindent\textbf{Zero-shot Generalization.}~
Under the single-degradation training setting, the standard-size \method~ achieves higher zero-shot performance than all compared universal methods on the real-world benchmarks (Table~\ref{table:zero_shot_generalization}).
For models trained on the complete CDD-50 degradation space, the performance further improves consistently as model size increases from $154$M to $2.9$B (Table~\ref{table:scaling_up_ours_average}), demonstrating the benefit of hierarchical scaling for zero-shot generalization.

\begin{figure}[!t]
	\centering
	\includegraphics[width=\columnwidth]{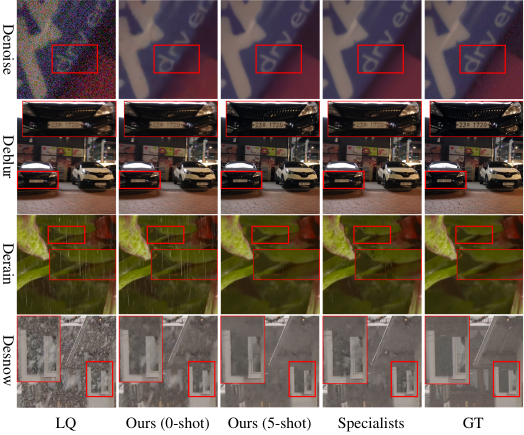}
	\caption{Visual comparison of zero-shot and five-shot \method~against specialist models. Five-shot fine-tuning improves degradation removal and detail recovery, as highlighted by the \textcolor{red}{red boxes}.}
	\label{fig:scaling_up_visual_comparison_specialist}
\end{figure}

\noindent\textbf{Few-shot Transfer.}~
For evaluation on one- and five-shot performance, we sample the whole one and five images, respectively, from the training sets of each task. 
Starting from the pretrained $2.9$B model, we freeze the shared backbone and fine-tune only the routers and LoRA experts. We keep the training settings as described in Section~\ref{sec:details}.
We repeat each experiment five times and report the mean and standard deviation, as results in Tables~\ref{table:scaling_up_compare_to_specialists} and~\ref{table:foundir}. The visual comparisons are in Fig.~\ref{fig:scaling_up_visual_comparison_specialist}.
Finally, five-shot fine-tuning exceeds the specialist methods on $14$ of $17$ test sets. 
Compared with zero-shot inference, few-shot adaptation improves more on the target distribution, which differs more from the pretraining distribution (\ie, realrain).

\noindent\textbf{Results and Analysis.}~ 
Across the evaluated model sizes, hierarchical scaling consistently improves both restoration accuracy and generalization. Multi-granularity routing alleviates the effect of degradation-estimation errors and allows uncertain inputs to use broader experts while retaining fine-grained specialization.

\section{Ablation Study and Analysis}
\label{sec:ablation_study}
We use a subset of the developed degradation space to ablate DR fineness, the multi-granularity degradation set, granularity-estimation-based routing, and the training losses. The synthetic degradation parameters are controlled to form \textit{in-dist.} and \textit{out-of-dist.} evaluation sets. Separately, we ablate the dense DR map and region-aware expert routing in the seven-task single-degradation setting.

\begin{table}[!t]
\caption{Ablation of DR fineness and the multi-granularity degradation set. Average PSNR (dB) is reported under \textit{in-dist.} and \textit{out-of-dist.} degradation parameters.}
\label{table:ablation_study}
\centering
\setlength{\tabcolsep}{3.5mm}
\resizebox{\columnwidth}{!}{%
\begin{tabular}{c | c  c  c  c }
\toprule

     \multirow{2}{*}{Routers}
  &  \multirow{2}{*}{\# Levels}
  &  \multirow{2}{*}{$\{$ \# DR Groups  $\}$} 
  &  \multicolumn{2}{c}{PSNR (dB)}    
  \\
  
  &
  &
  & \textit{In-dist.} 
  & \textit{Out-of-dist.}   
  \\
  
\midrule
 \multirow{5}{*}{DER}  & 1 & $\{$ 1 $\}$          & 22.06 & 17.23  \\
& 1 & $\{$ 2 $\}$          & 22.42 & 17.51  \\
& 1 & $\{$ 4 $\}$          & 23.75 & 18.49  \\
& 1 & $\{$ 8 $\}$          & 24.22 & 18.35  \\
& 1 & $\{$ 16 $\}$         & 24.30 & 18.44  \\
\midrule
\multirow{4}{*}{DER, GER}& 2 & $\{$ 1, 8 $\}$       & 24.27 & 18.86  \\
& 2 & $\{$ 4, 8 $\}$       & 24.44 & 19.06  \\
& 3 & $\{$ 1, 4, 8 $\}$    & 24.46 & 19.45  \\
& 4 & $\{$ 1, 4, 8, 16$\}$ & 24.41 & 19.47  \\
\bottomrule
\end{tabular}
}
\end{table}

\begin{figure}[!t]
	\centering
    \includegraphics[width=0.85\columnwidth]{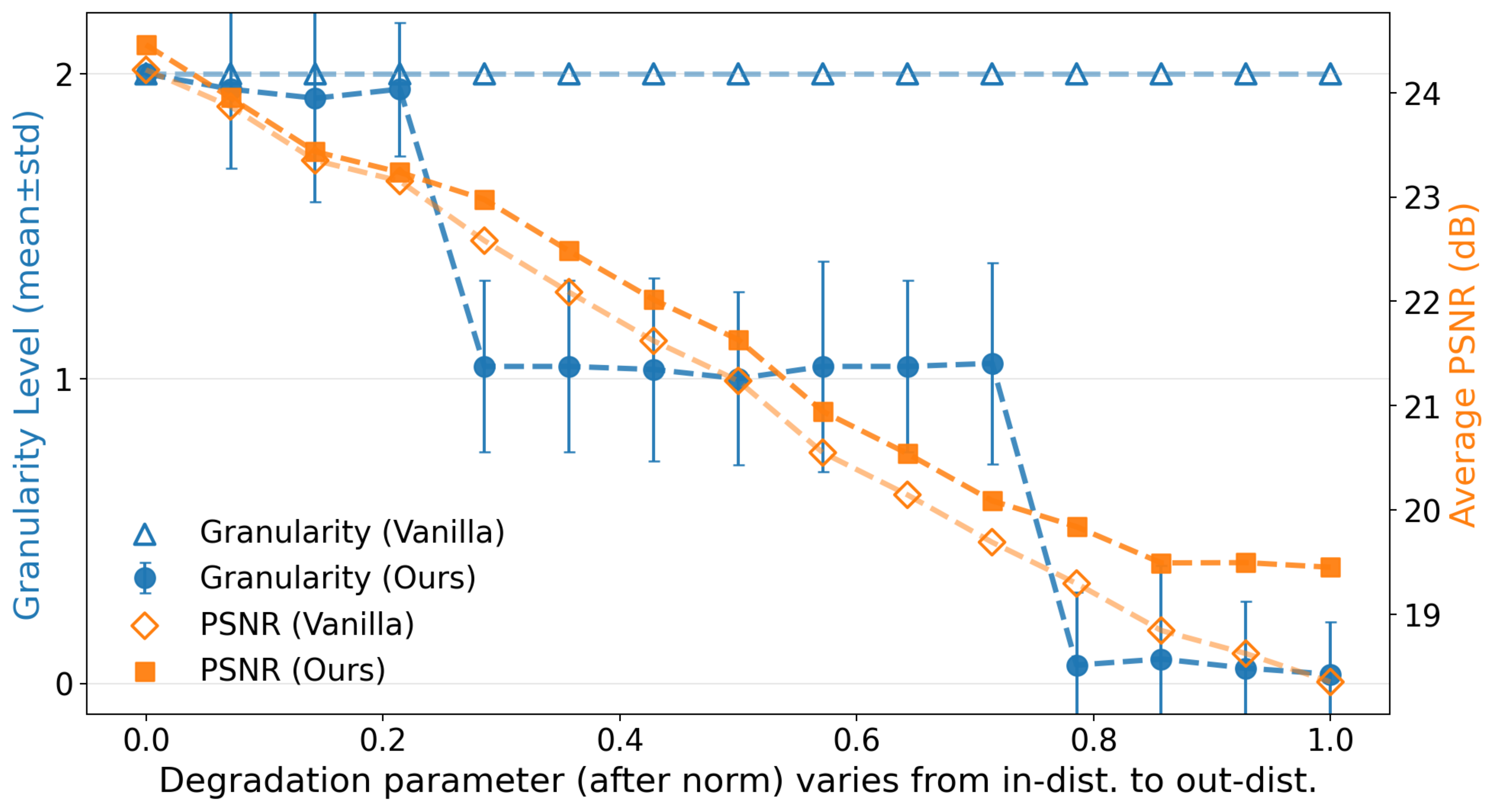} 
    \caption{Average selected hierarchy level and corresponding PSNR for the full and \textit{Vanilla} variants as degradation parameters shift from \textit{in-dist.} to \textit{out-of-dist.}}
    \label{fig:routing}
\end{figure}

\begin{table}[!t]
\caption{Ablation of the dense DR map and region-aware expert routing. PSNR (dB) is reported per task, and Average is the mean over the seven tasks.}
\label{table:spatial_variant}
\centering
\footnotesize
\setlength{\tabcolsep}{2pt}
\renewcommand{\arraystretch}{1.12}
\begin{tabular*}{\columnwidth}{@{\extracolsep{\fill}}lcccccccc@{}}
\toprule%
Variant
& Rain
& Haze
& Noise
& Blur
& Low-light
& Snow
& CAR
& Avg. \\
\midrule%
Baseline & 40.22 & 34.94 & 31.39 & 30.86 & 24.42 & 30.53 & 30.31 & 31.81 \\
+ Dense DR & 40.31 & 35.09 & 31.44 & 31.02 & 24.52 & 30.59 & 31.27 & 32.03 \\
\makecell[l]{+ Region-Aware\\Expert Routing} & 40.35 & 35.51 & 31.48 & 31.59 & 24.64 & 30.66 & 31.48 & 32.24 \\
\bottomrule%
\end{tabular*}
\end{table}

\begin{table}[!t]
\caption{Ablation of the loss functions.}
\label{table:loss_functions}
\centering
\setlength{\tabcolsep}{5mm}
\resizebox{\columnwidth}{!}{%
\begin{tabular}{c c c c c }
\toprule%
 \multirow{2}{*}{$\ell_1$} & \multirow{2}{*}{$\mathcal{L}_{\mathrm{dg}}$} & \multirow{2}{*}{$\mathcal{L}_{\mathrm{aux}}$} & \multicolumn{2}{c}{PSNR (dB)}     \\
           &                      &                      & \textit{In-dist.} & \textit{Out-of-dist.} \\
\midrule%
 \cmark & \xmark  & \xmark     &  24.14         & 18.37          \\
 \cmark & \cmark  & \xmark           &  24.35         & 18.76          \\
 \cmark & \cmark  & \cmark &  24.46         & 19.45  \\
\bottomrule%
\end{tabular}
}
\end{table}

\subsection{Effect of the Multi-Granularity Degradation Modeling}

\noindent\textbf{Effect of DR Fineness.}~
The upper part of Table~\ref{table:ablation_study} evaluates a single DER with different numbers of DR groups.
Increasing the number of groups from $1$ to $8$ improves \textit{in-dist.} PSNR from $22.06$ to $24.22$ dB, showing the benefit of finer specialization.
Increasing the number further to $16$ yields only $0.08$ dB additional improvement. In contrast, \textit{out-of-dist.} performance is highest with $4$ groups ($18.49$ dB) among the single-granularity variants and does not improve as the partition becomes finer. Thus, a single fixed granularity cannot simultaneously satisfy \textit{in-dist.} specialization and \textit{out-of-dist.} generalization.

\noindent\textbf{Effect of Different Granularities.}~
The lower part of Table~\ref{table:ablation_study} evaluates multi-granularity configurations using both DER and GER. The three-level configuration $\{1,4,8\}$ achieves $24.46$ dB on \textit{in-dist.} data and $19.45$ dB on \textit{out-of-dist.} data, outperforming all single-granularity variants on both evaluation settings. Adding a fourth level with $16$ groups changes the results to $24.41$ and $19.47$ dB, providing only a $0.02$ dB \textit{out-of-dist.} gain while reducing \textit{in-dist.} performance and requiring more experts. 
We therefore use $\{1,4,8\}$ for the experiments in the triple-degradation setting.

\subsection{Effect and Analysis of Multi-Granularity Expert Routing}

\noindent\textbf{Effect of Granularity-Estimation-Based Routing.}~
To isolate GER, we compare the full model with a \textit{Vanilla} variant that always selects the finest hierarchy level. We normalize the degradation parameter range from \textit{in-dist.} to \textit{out-of-dist.} into $[0,1]$, uniformly sample $15$ points, and apply the corresponding degradations to Urban100. Fig.~\ref{fig:routing} reports the average selected hierarchy level and PSNR at each point. As the degradation parameters move away from the training distribution, GER increasingly selects coarser hierarchy levels, whereas \textit{Vanilla} remains restricted to the finest level. This adaptation preserves \textit{in-dist.} accuracy while improving robustness under larger distribution shifts.

\noindent\textbf{Visualization of Routing Results.}~ 
\begin{figure}[t!]
	\centering
	\includegraphics[width=\columnwidth]{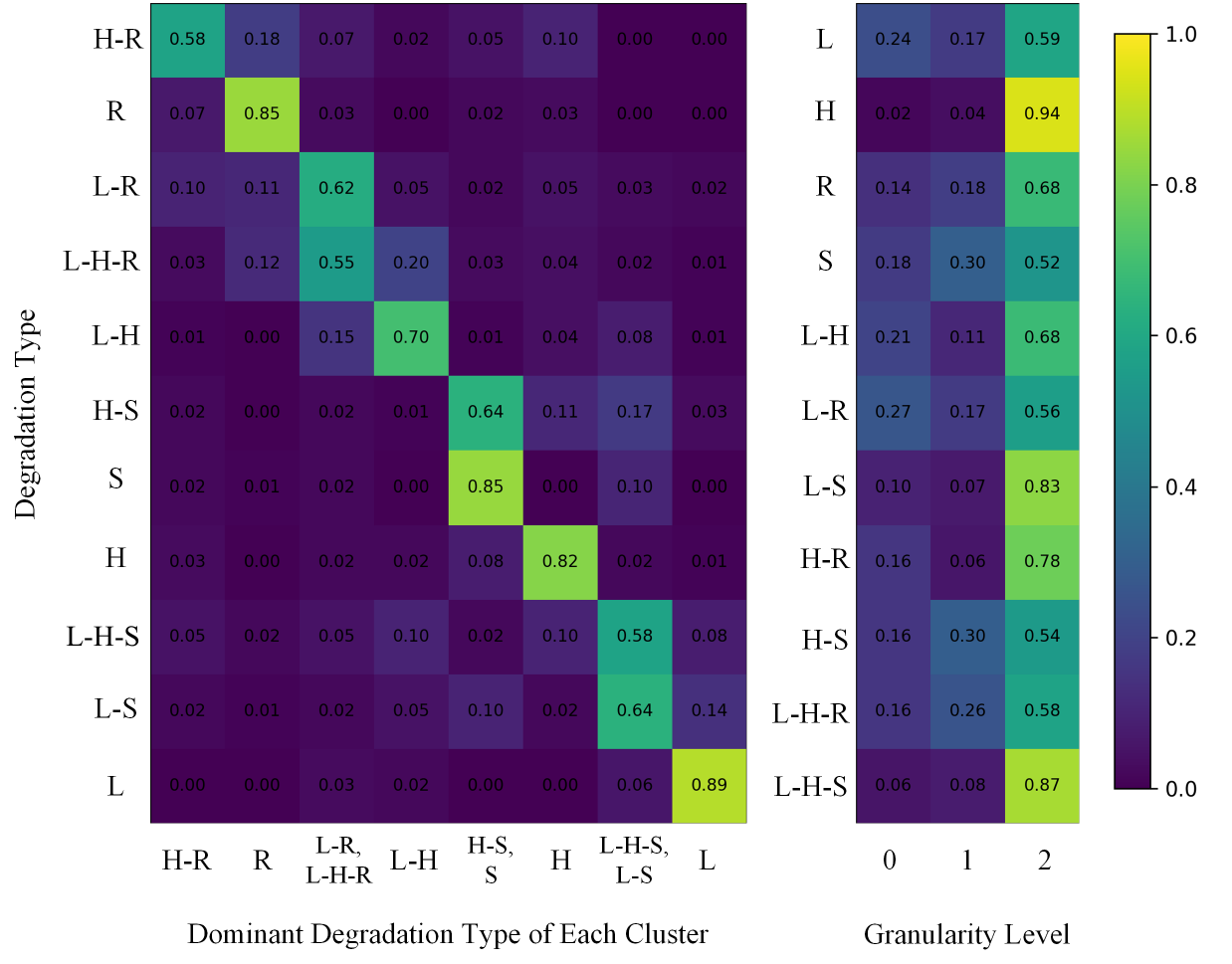}
	\caption{Routing frequencies on CDD-11. Left: degradation-estimation-based router (DER) assignments to the eight finest-level DR groups, whose dominant degradation types are shown on the horizontal axis. Right: granularity-estimation-based router (GER) assignments to the hierarchy levels. Rows correspond to the 11 CDD-11 degradation subsets.}
	\label{fig:routing_confidence}
\end{figure}   

We inspect the finest-level DR grouping and expert assignments on CDD-11. The dominant degradation types of the eight finest-level DR groups are:
(1) \{Haze+Rain (H+R)\},
(2) \{Rain (R)\},
(3) \{Low-light+Rain (L+R), Low-light+Haze+Rain (L+H+R)\},
(4) \{Low-light+Haze (L+H)\},
(5) \{Haze+Snow (H+S), Snow (S)\},
(6) \{Haze (H)\},
(7) \{Low-light+Haze+Snow (L+H+S), Low-light+Snow (L+S)\},
(8) \{Low-light (L)\}.
Fig.~\ref{fig:routing_confidence} shows different finest-level group assignments and granularity preferences across degradation subsets. GER selects both coarse and fine levels, illustrating the use of restoration capacity at multiple granularities.

\subsection{Effect of Dense DR and Region-Aware Expert Routing}

\begin{figure}[!t]
    \centering
    \includegraphics[width=\columnwidth]{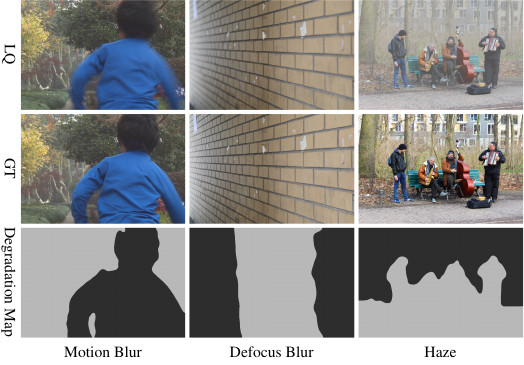}
    \caption{Visualization of the dense DR map for real-world spatially varying degradations, \ie, motion blur (ReLoBlur~\cite{reloblur}), defocus blur (DPDD~\cite{dpdd}), and haze (FoundIR~\cite{foundir}). Gray and black indicate relatively lower and higher degradation severity, respectively.}
    \label{fig:dense_dr_map_vis}
\end{figure}

Table~\ref{table:spatial_variant} ablates the dense DR map and region-aware expert routing. The first row is the UniRestorer baseline, which uses an image-level DR for global expert routing. 
The second row replaces the image-level DR with the dense DR map but spatially averages it before expert routing, increasing the seven-task average PSNR from $31.81$ to $32.03$ dB. 
The third row preserves the spatial information in the dense DR map for region-aware expert routing, increasing the average to $32.24$ dB. 
On the two representative spatially varying degradations, our design improves dehazing from $34.94$ to $35.51$ dB and deblurring from $30.86$ to $31.59$ dB. These gains support the use of the dense DR map and region-aware expert routing.
Fig.~\ref{fig:dense_dr_map_vis} visualizes the responses of the dense DR map on motion blur, defocus blur, and haze, illustrating spatially varying responses to these degradations.

\subsection{Effect of Loss Functions}

Table~\ref{table:loss_functions} evaluates the loss functions used during training.
Compared with the $\ell_1$ restoration objective alone, adding $\mathcal{L}_{\mathrm{dg}}$ improves PSNR from $24.14$ to $24.35$ dB on the \textit{in-dist.} data and from $18.37$ to $18.76$ dB on the \textit{out-of-dist.} data.
Adding the GER load-balancing loss $\mathcal{L}_{\mathrm{aux}}$ increases PSNR to $24.46$ dB on the \textit{in-dist.} data and $19.45$ dB on the \textit{out-of-dist.} data. 

\section{Conclusion and Limitations}
\label{sec:conclusion}
In this paper, we presented \method, a universal image restoration framework that scales capacity through hierarchical degradation-specific experts.
Dense degradation representations and multi-granularity routing enable spatially adaptive expert selection while reducing sensitivity to degradation-estimation errors.
We constructed CDD-50 and evaluated six variants from $154$M to $2.9$B parameters, observing approximately log-linear PSNR trends within this range.
Our method outperforms the compared universal restoration methods and demonstrates zero-shot generalization and few-shot transfer, surpassing the compared specialist models on $14$ of $17$ test sets after five-shot fine-tuning.
These results support hierarchical capacity expansion as a promising approach to scalable universal restoration.

A limitation is that our degradation hierarchy and expert pool are constructed offline and remain fixed during inference.
Inspired by recent work on self-improving agents~\cite{zhang2025darwin}, a future direction is to extend this modular framework toward a recursively self-improving (RSI) restoration agent.
Such an agent could diagnose failures, refine degradation groups, add or update experts, and revise their routing, with proposed changes assessed on held-out data.
By learning from the outcomes of its own updates, such an agent could improve both restoration performance and its strategy for refining and expanding the expert hierarchy.
This would extend our hierarchy model scaling strategy from predefined capacity expansion toward feedback-driven improvement, allowing restoration capabilities to evolve with changing real-world degradations.

\bibliographystyle{IEEEtran}
\bibliography{reference}

\end{document}